\documentclass[twocolumn]{article}
\usepackage{iftex}
\ifPDFTeX\pdfoutput=1\pdfminorversion=7\fi
\usepackage{arxiv}
\AtBeginDocument{%
  \newgeometry{margin=1in,headheight=14pt,headsep=25pt,footskip=30pt}%
}
\usepackage[utf8]{inputenc}
\usepackage[T1]{fontenc}
\usepackage{amsmath,amssymb,latexsym}
\usepackage{microtype}
\usepackage{inconsolata}
\usepackage{graphicx}
\usepackage{xcolor}
\usepackage{booktabs,tabularx,array,longtable,colortbl}
\usepackage{fvextra}
\usepackage{caption}
\usepackage[authoryear,round]{natbib}
\renewcommand{\cite}{\citep}
\usepackage{xurl}
\definecolor{linkblue}{RGB}{26,65,112}
\usepackage[colorlinks=true,allcolors=linkblue,breaklinks=true]{hyperref}
\renewcommand{\shorttitle}{CD-OPD}
\hypersetup{
  pdftitle={CD-OPD: Contextual Digest for On-Policy Distillation},
  pdfauthor={Mingyuan Li, Yucheng Li, Xiaxia Wang, Gong Cheng},
  pdfsubject={On-policy knowledge distillation},
  pdfkeywords={on-policy distillation, contextual digest, language models}
}

\definecolor{oursbg}{RGB}{232,242,252}
\definecolor{improvementbg}{RGB}{240,247,252}
\definecolor{caseblue}{HTML}{245C92}
\definecolor{casered}{HTML}{A42B26}
\definecolor{casegreen}{HTML}{226746}
\definecolor{casegray}{HTML}{626262}
\newcommand{\casekey}[1]{{\color{caseblue}\boldsymbol{#1}}}
\newcommand{\casewrong}[1]{{\color{casered}\boldsymbol{#1}}}
\newcommand{\caseright}[1]{{\color{casegreen}\boldsymbol{#1}}}
\newcommand{\casequote}[1]{``#1''\par}
\newcommand{\casecell}[2]{%
  \begin{minipage}[t]{\linewidth}
  \vspace{1pt}
  \setlength{\parindent}{0pt}
  \setlength{\parskip}{2.5pt}
  \setlength{\abovedisplayskip}{4pt}
  \setlength{\belowdisplayskip}{4pt}
  \setlength{\abovedisplayshortskip}{3pt}
  \setlength{\belowdisplayshortskip}{3pt}
  \setlength{\jot}{2pt}
  \textbf{#1}\par\vspace{2pt}
  #2\par\vspace{4pt}
  \end{minipage}%
}

\usepackage{tcolorbox}
\tcbuselibrary{listings,breakable}
\newtcblisting{cdprompt}[1]{%
  listing only,breakable,
  title={#1},fonttitle=\sffamily\bfseries\small,
  colback=black!2,colframe=black!25,
  colbacktitle=black!8,coltitle=black,
  boxrule=0.4pt,sharp corners,
  boxsep=1mm,left=2mm,right=2mm,top=1mm,bottom=1mm,
  before skip=6pt,after skip=8pt,
  listing options={
    language={},basicstyle=\sffamily\footnotesize,
    columns=fullflexible,keepspaces=true,
    breaklines=true,breakatwhitespace=true,
    breakautoindent=false,breakindent=0pt,
    showstringspaces=false
  }
}

\title{CD-OPD: Contextual Digest for On-Policy Distillation}
\author{%
  Mingyuan Li\textsuperscript{1}\quad
  Yucheng Li\textsuperscript{1}\quad
  Xiaxia Wang\textsuperscript{2}\quad
  Gong Cheng\textsuperscript{1,*}\\[5pt]
  {\small \textsuperscript{1}State Key Laboratory of Novel Software Technology,}\\[-1pt]
  {\small Nanjing University, Nanjing, China}\\[2pt]
  {\small \textsuperscript{2}Department of Computer Science, University of Oxford}\\[4pt]
  {\small \textsuperscript{*}Corresponding author:
    \href{mailto:gcheng@nju.edu.cn}{\texttt{gcheng@nju.edu.cn}}}%
}

\date{}

\begin{document}
\ifXeTeX\fi
\twocolumn[{\begin{@twocolumnfalse}
\maketitle
\end{@twocolumnfalse}}]

\begin{abstract}
Teacher takeover in on-policy distillation (OPD) can produce incorrect responses even when the student and teacher independently answer the same problem correctly, a failure we quantify using Collaboration Failure Rate (CFR).
To address it, we propose Contextual Digest for On-Policy Distillation (CD-OPD).
CD-OPD structures the earlier context of teacher guidance through a teacher-generated dependency graph, curating the required problem conditions and preceding reasoning as node summaries alongside the unchanged guidance in a Contextual Digest before student continuation.
We further develop online meta-reweighting, which compares Digest segment gradients with a task gradient from independent student responses to improve student-only performance, favoring updates predicted to reduce task loss.
Across eight benchmarks, CD-OPD improves average accuracy by 1.32--7.05 percentage points over representative existing OPD methods.
It also reduces CFR by 27.4\%--56.4\% relative to existing takeover-based OPD methods.
\end{abstract}

\section{Introduction}

Knowledge distillation (KD) trains a smaller student model with supervision from a larger teacher model to reduce inference costs~\cite{KD}.
For language generation and reasoning, teacher-generated responses and rationales can serve as training targets~\cite{SequenceKD,hsieh-etal-2023-distilling}.
To reduce the distribution mismatch between teacher-generated training responses and student-generated responses at inference time, on-policy distillation (OPD) trains on responses sampled from the current student policy.
At each token position, the teacher takes the problem and preceding student-generated tokens as input and provides its next-token probability
distribution as supervision~\cite{opd1,opd2,opd3}.
\begin{figure}[t]
    \centering
    \includegraphics[width=\linewidth]{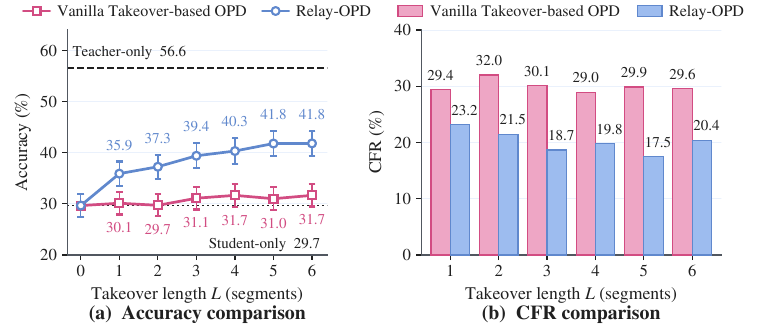}
    \caption{Accuracy and CFR under teacher takeover (1,024 mathematical problems randomly sampled from our local copy of DAPO-Math-17k-dedup; student: Qwen3-1.7B; teacher: Qwen3-4B-Instruct-2507; accuracy reported as mean@4). }
    \label{Figure1}
\end{figure}

\begin{figure*}[!t]
    \centering
    \includegraphics[
    width=1.0\textwidth,
    trim=0 100 0 0,
    clip
    ]{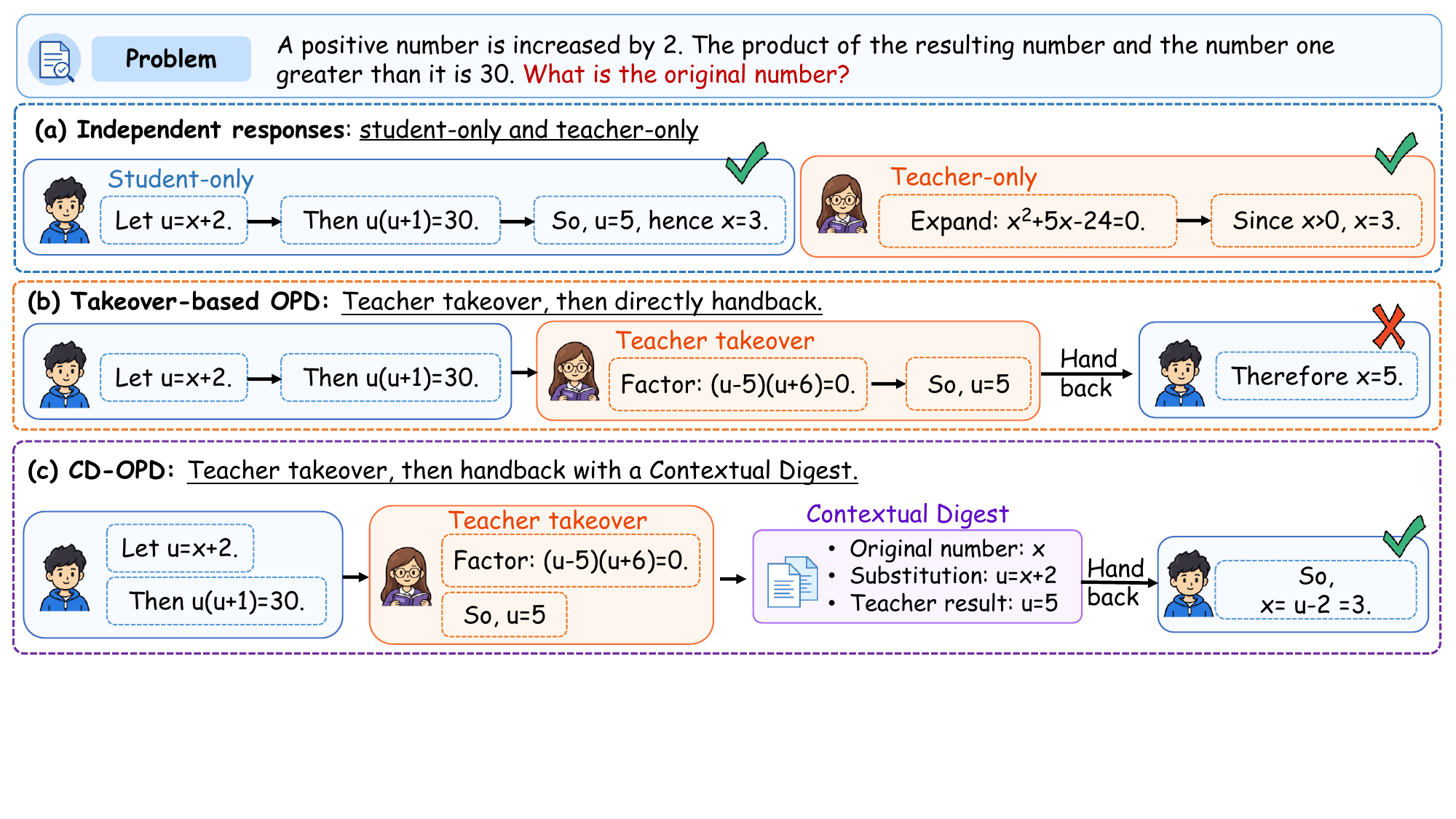}
\caption{
Comparison of independent student and teacher responses, takeover-based OPD, and CD-OPD.
}
    \label{Figure2}
\end{figure*}
However, such a teacher evaluates the sampled response without revising it, so its supervision can remain conditioned on prefixes containing earlier reasoning errors~\cite{rethinking_opd, fu2026revisiting,fastopd}.
To provide corrections during generation,
teacher takeover allows the teacher to
generate a continuation from a student-generated prefix
before the student resumes; the combined response
is used for distillation~\cite{switch,relay-opd}.
A key design choice in teacher takeover is when to intervene.
We consider two representative strategies.
Fixed-prefix takeover begins at a predetermined student-prefix length~\cite{rethinking_opd}.
Reflection-triggered takeover instead compares the teacher's and student's next-token predictions at the same prefix.
The teacher takes over when its most probable next token is a reflection token (e.g., \textit{Wait}), while the student's top-$K$ predictions contain no reflection tokens~\cite{relay-opd}.
Both strategies determine when the teacher intervenes, but the final response also depends on \textit{how the student continues after teacher guidance}; if the student misapplies the guidance, the intervention may be ineffective or even misleading.
We observe incorrect takeover responses even when the student and teacher independently answer the same problem correctly.
Overall accuracy gains may obscure these failures.

To uncover this failure mode, we define \textbf{Collaboration Failure Rate (CFR)} as the proportion of incorrect takeover responses among cases where the corresponding student-only and teacher-only responses are both correct.
Figure~\ref{Figure1} compares accuracy and CFR for  fixed-prefix takeover (Vanilla Takeover-based OPD) and reflection-triggered takeover (Relay-OPD), with the takeover length $L$ ranging from one to six teacher-generated segments.
Across these lengths, Relay-OPD achieves higher accuracy and lower CFR than Vanilla Takeover-based OPD,
yet its CFR of 17.5\%--23.2\% corresponds to roughly one counterproductive takeover response in every five cases with correct student-only and teacher-only responses.
Increasing the takeover length does not consistently reduce CFR: from $L=5$ to $L=6$, Relay-OPD's reported accuracy remains at 41.8\%, while CFR rises from 17.5\% to 20.4\%.
\textit{These findings motivate assessing collaboration failures separately from overall accuracy and examining how the student effectively uses teacher guidance during continuation.}
Figure~\ref{Figure2} illustrates a collaboration failure of the kind measured by CFR.
The problem asks for the original number $x$.
Both the student-only and teacher-only responses correctly obtain $x=3$ (Figure~\ref{Figure2}(a)).
Under takeover-based OPD, the student prefix defines $u=x+2$ and expresses the given condition as $u(u+1)=30$.
The teacher factors the equation as $(u-5)(u+6)=0$ and correctly obtains $u=5$ (Figure~\ref{Figure2}(b)).
The student then mistakenly reports $x=5$, treating the value of $u$ as the original number.
This example shows a failure to use existing context despite correct teacher guidance: the student does not apply the definition $u=x+2$, already recorded in its prefix, to recover $x$ from the teacher's result.

To address this failure mode, we propose
\emph{\textbf{C}ontextual \textbf{D}igest for \textbf{O}n-\textbf{P}olicy \textbf{D}istillation
(\textbf{CD-OPD})}. 
A teacher-generated dependency graph selects problem conditions and preceding reasoning for summarization under the current guidance in a Contextual Digest.
The Digest incorporates the teacher's corrections and is inserted after the guidance, before the student resumes (Figure~\ref{Figure2}(c)).
A Digest segment that aids continuation need not improve independent reasoning when emphasized during training.
We therefore develop online meta-reweighting to favor Digest-segment OPD updates predicted to reduce student-only task loss.
During training, we determine segment weights by comparing segment gradients from takeover-based responses with a task gradient from separate student-only training responses.

Our \textbf{contributions} are twofold.
\begin{itemize}
    \item 
    We introduce CFR to measure the rate of incorrect takeover responses among cases where both the student-only and teacher-only responses are correct. 
    Our analysis reveals collaboration failures that persist despite the overall accuracy gains of teacher takeover.
    \item 
    We propose CD-OPD, which uses a teacher-generated dependency graph to condense the existing context and current teacher guidance into a Contextual Digest before student continuation, helping the student apply the guidance using relevant earlier context.
    Online meta-reweighting uses the task gradient and segment gradients to weight each Digest segment's OPD loss, favoring updates expected to improve student-only performance.
\end{itemize}

Our code is publicly available at \url{https://anonymous.4open.science/r/CD-OPD-291D/}.

\section{Related Work}
\paragraph{On-Policy Distillation.}
On-policy distillation trains the student on responses sampled from its current policy, with token-level teacher supervision conditioned on the resulting prefixes~\cite{GDK}. 
MiniLLM~\cite{MiniLLM} uses policy-gradient updates to minimize reverse KL divergence between the student and teacher distributions.
OPD+~\cite{opd3} corrects advantage estimates for general \(f\)-divergence objectives by accounting for the reward's dependence on student parameters.
Teacher supervision can nevertheless remain conditioned on prefixes containing student reasoning errors.
On long student-generated responses, Rethinking OPD~\cite{rethinking_opd} reports decreasing accuracy gains from teacher continuation as the retained prefix length increases.
FastOPD~\cite{fastopd} and Early Stopping Rollout~\cite{opd_esr} reduce rollout cost by terminating student generation early and restricting distillation to the retained prefixes.
\textit{Refining policy updates or shortening student rollouts does not by itself provide corrective guidance within the sampled response.
Takeover-based OPD complements these approaches by allowing the teacher to generate corrective continuations from student prefixes before the student resumes.}

\paragraph{Takeover-based OPD.}
Teacher intervention modifies student-generated responses through revision or corrective continuation.
TRD~\cite{trd} asks the teacher to revise student-generated responses and trains the student on the revised responses through distillation.
At the token level, Speculative Knowledge Distillation~\cite{SKD} replaces student-proposed tokens outside the teacher's top-\(K\) predictions with tokens sampled from the teacher.
SWITCH~\cite{switch} determines whether the teacher or student generates each token according to the Jensen--Shannon divergence between their next-token distributions.
MOTAB~\cite{motab} generates a teacher continuation after backtracking to an earlier student prefix, then appends it to the interruption point for fine-tuning.
Relay-OPD~\cite{relay-opd} instead triggers teacher takeovers at the current prefix based on teacher--student disagreement over reflection tokens; the student resumes generation after the teacher provides guidance.
\textit{
Existing takeover methods primarily control teacher intervention but leave the dependencies between teacher guidance and earlier context implicit, so correct guidance may still be misapplied.
Our proposed CD-OPD explicitly uses a teacher-generated dependency graph to identify the required problem conditions and preceding reasoning, which are summarized under the current guidance as an enrichment, helping the student apply the guidance using the relevant earlier context during continuation.
} 
\begin{figure*}[t]
    \centering
    \includegraphics[
    width=1.0\textwidth,
    trim=0 140 0 0,
    clip
]{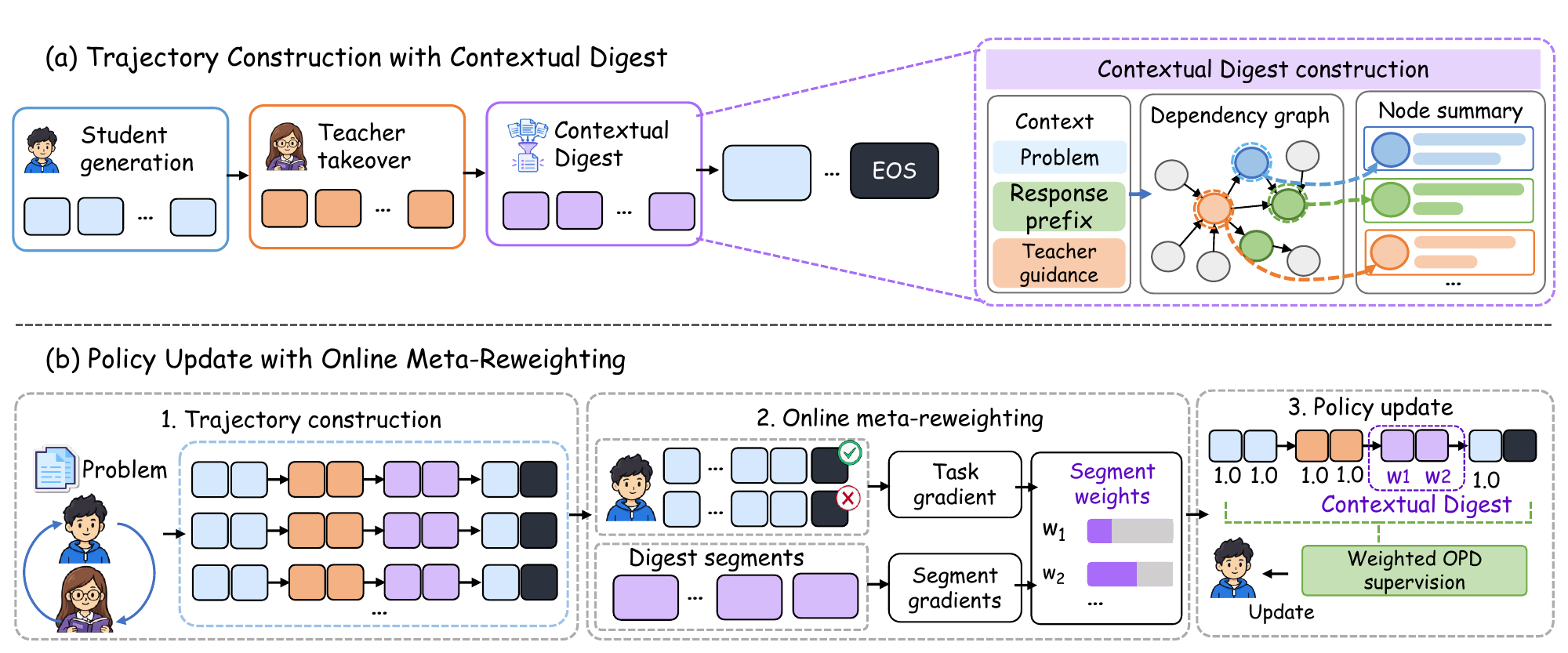}
    \caption{
Overview of CD-OPD:
(a) Trajectory Construction with Contextual Digest
(Section~\ref{sec:trajectory});
(b) Policy Update with Online Meta-Reweighting
(Section~\ref{sec:policy_update}).
}
    \label{Figure3}
\end{figure*}

\section{Method}
\subsection{Preliminaries}
\paragraph{On-Policy Distillation.} 
Let $\pi_\theta$ denote the student policy and $\pi_T$ the frozen teacher policy, with a shared token vocabulary $\mathcal V$.
Given a training problem $x$, OPD samples an $N$-token response $y=(y_1,\ldots,y_N)$ from $\pi_\theta$.
For $t=1,\ldots,N$, the prefix $h_t=(x,y_{<t})$ contains the problem and preceding student-generated tokens.
For $v\in\mathcal V$, the student and teacher next-token probabilities conditioned on the same prefix $h_t$ are
\begin{equation}
p_t(v)=\pi_\theta(v\mid h_t),
\qquad
q_t(v)=\pi_T(v\mid h_t).
\end{equation}
At each fixed prefix $h_t$, we use the reverse Kullback--Leibler (KL) divergence~\cite{kullback1951information,MiniLLM}:
\begin{equation}
    D_{\mathrm{KL}}(p_t\|q_t)
=\sum_{v\in\mathcal V}
p_t(v)\log\frac{p_t(v)}{q_t(v)}.
\end{equation}
We minimize this divergence averaged over response-token positions.
For a student-sampled token $y_t\sim p_t$, the token-level OPD advantage is
\begin{equation}
A_t^{\mathrm{OPD}}
=
\log q_t(y_t)-\log p_t(y_t).
\label{eq:opd_advantage}
\end{equation}
We update only $\theta$ by policy-gradient ascent, weighting $\nabla_\theta\log p_t(y_t)$ by $A_t^{\mathrm{OPD}}$.
The sampled prefixes $h_t$ and advantages $A_t^{\mathrm{OPD}}$ are treated as constants during the update.
Problem and padding tokens are excluded from the loss.
The next iteration samples responses from the updated student policy $\pi_\theta$.

\paragraph{Teacher Takeover.}
During response generation, the teacher may take over at selected prefixes~\cite{switch,motab}.
Let $z=(z_1,\ldots,z_{N_z})$ denote the resulting response, and let $h_t^z=(x,z_{<t})$ denote its prefix at position $t$.

Following Relay-OPD~\cite{relay-opd}, we determine takeover positions during student generation using a predefined reflection-token set $\mathcal R\subseteq\mathcal V$ (e.g., token variants of \textit{Wait}, \textit{But}, and \textit{However}). 
Given $h_t^z$, let $a_t^T$ denote the teacher's
most probable next token and $\mathcal K_t$ the
set of the student's $K$ most probable next tokens:
\begin{equation}
\begin{aligned}
a_t^T
&=
\arg\max_{v\in\mathcal V}
\pi_T(v\mid h_t^z),\\
\mathcal K_t
&=
\operatorname*{arg\,topK}_{v\in\mathcal V}
\pi_\theta(v\mid h_t^z).
\end{aligned}
\label{eq:takeover_predictions}
\end{equation}
Here, $\operatorname{arg\,topK}$ returns the set of $K$ tokens with the highest probabilities.
The takeover criterion is
\begin{equation}
\phi(h_t^z)
=
\mathbf{1}[a_t^T\in\mathcal R]
\cdot
\mathbf{1}[\mathcal K_t\cap\mathcal R=\varnothing],
\label{eq:takeover_trigger}
\end{equation}
where $\mathbf{1}[\cdot]$ denotes the indicator function.
We allow at most $M$ takeovers per response and $L$ teacher-generated segments per takeover.
Each teacher-generated segment is a paragraph delimited by a blank line.
When $\phi(h_t^z)=1$ and fewer than $M$ takeovers have occurred, the teacher generates the current guidance $T$, beginning with $a_t^T$.
Otherwise, the next token is sampled from $\pi_\theta$.
After each takeover, the student resumes from the extended prefix if the response is unfinished.
The completed response $z$ is used for token-level distillation, with both policies evaluated at each prefix $h_t^z$.

\subsection{Overview}
As shown in Figure~\ref{Figure3}, CD-OPD operates at two stages of takeover-based OPD: trajectory construction and policy update.
After each takeover, the teacher constructs a dependency graph from the context comprising the problem, the response prefix before takeover, and the current teacher guidance.
The graph identifies the problem conditions and preceding reasoning needed to apply the guidance in answering the problem.
The teacher summarizes this selected information under the current guidance, producing node summaries that are combined with the unchanged guidance in a Contextual Digest.
The Digest is appended after the guidance for student continuation (Figure~\ref{Figure3}(a)).
During policy updates, online meta-reweighting uses a task gradient from student-only responses and segment gradients to determine each Digest segment's OPD loss weight, favoring updates expected to improve student-only performance (Figure~\ref{Figure3}(b)).

\subsection{Trajectory Construction with Contextual Digest}
\label{sec:trajectory}
At takeover position $t$, the current teacher guidance $T$ may depend on problem conditions in $x$ and preceding reasoning in $z_{<t}$ without restating them.
CD-OPD appends a Contextual Digest $D$ after $T$ to help the student apply the guidance using this earlier information.

The teacher constructs a directed dependency graph $\mathcal G=(\mathcal N,\mathcal E)$ from the context $(x,z_{<t},T)$.
Each node $n_i\in\mathcal N$ represents one condition, variable definition, intermediate result, or reasoning step explicitly stated in the inputs.
Its source spans are the text spans in $x$, $z_{<t}$, or $T$ that express its content.
Source spans share a node only when their content, variable meanings, units, or conditions overlap.
The teacher adds an edge $n_i\to n_j$ to $\mathcal E$ when it identifies information in $n_i$ needed to interpret or apply the content of $n_j$ in answering the problem.
If $T$ explicitly corrects a statement in $z_{<t}$, the graph uses the correction in place of that statement and cites its source span in $T$; $z_{<t}$ remains unchanged.

The nodes associated with at least one source span in $T$ form $\mathcal N_T\subseteq\mathcal N$.
Starting from $\mathcal N_T$, we recursively follow incoming edges to select
\begin{equation}
\mathcal N_D
=
\mathcal N_T
\cup
\operatorname{Anc}_{\mathcal G}(\mathcal N_T),
\label{eq:digest_nodes}
\end{equation}
where $\operatorname{Anc}_{\mathcal G}(\mathcal N_T)$ contains all nodes with a directed path to a node in $\mathcal N_T$.
We topologically sort the nodes in $\mathcal N_D$ according to their dependencies in $\mathcal G$.
Let $J=|\mathcal N_D\setminus\mathcal N_T|+1$.
For each node in $\mathcal N_D\setminus\mathcal N_T$, the teacher summarizes its source spans under the current guidance $T$.
The resulting node summaries form the Digest segments $D_1,\ldots,D_{J-1}$ in this order.
We copy $T$ unchanged as the final Digest segment, $D_J=T$.
The resulting Contextual Digest is
\begin{equation}
D=D_1\oplus\cdots\oplus D_J,
\label{eq:digest_construction}
\end{equation}
where $\oplus$ denotes text concatenation.
If $\mathcal N_D=\mathcal N_T$, then $D=T$.
If the response is unfinished, the student resumes generation conditioned on $(x,z_{<t},T,D)$.
The resulting response, including the inserted Digest tokens, is still denoted by $z$.

\subsection{Policy Update with Online Meta-Reweighting}
\label{sec:policy_update}
A Digest segment may help the current continuation, but increasing its OPD loss weight need not improve student-only performance.
We therefore compare each segment's OPD gradient with a task gradient from student-only responses, assigning larger weights to updates predicted to reduce student-only task loss.

\paragraph{Trajectory Construction.}
At the start of each training iteration, we set $\bar\theta\leftarrow\theta$ and keep $\bar\theta$ fixed throughout the iteration.
All student generation uses $\pi_{\bar\theta}$, both in responses with Contextual Digest and in separate student-only responses.
For task feedback, we sample $B$ distinct training problems $\{\tilde x_b\}_{b=1}^{B}$ and generate $R\ge2$ responses $\{\tilde y^{(b,k)}\}_{k=1}^{R}$ per problem, without teacher takeover or Contextual Digest.

\paragraph{Online Meta-Reweighting.}
Let $r_{b,k}\in\{0,1\}$ indicate whether the final answer in $\tilde y^{(b,k)}$ is correct.
To obtain a task gradient that favors correct responses over incorrect alternatives to the same problem, we define the task advantage as
\begin{equation}
A_{b,k}^{\mathrm{task}}
=
r_{b,k}
-
\frac{1}{R-1}
\sum_{\substack{k'=1\\k'\ne k}}^{R}
r_{b,k'}.
\label{eq:task_advantage}
\end{equation}
Let $|\tilde y^{(b,k)}|$ denote the number of tokens in $\tilde y^{(b,k)}$.
We define the task loss and its gradient using the length-normalized log-probability assigned by $\pi_\theta$ to each sampled response, weighted by $A_{b,k}^{\mathrm{task}}$:
\begin{align}
\scalebox{0.92}{$\displaystyle
\mathcal L_{\mathrm{task}}(\theta)
$}
&\scalebox{0.92}{$\displaystyle
{}=
-\frac{1}{B\cdot R}
\sum_{b=1}^{B}\sum_{k=1}^{R}
A_{b,k}^{\mathrm{task}}
\frac{
\log\pi_\theta
(\tilde y^{(b,k)}\mid\tilde x_b)
}{
|\tilde y^{(b,k)}|
},
$}
\notag\\
g_{\mathrm{task}}
&=
\left.
\nabla_\theta\mathcal L_{\mathrm{task}}(\theta)
\right|_{\theta=\bar\theta}.
\label{eq:task_gradient}
\end{align}
The conditional log-probability of a response under $\pi_\theta$ is obtained by summing the token log-probabilities given the problem and preceding response tokens.
By favoring higher-scoring responses to each problem, $\mathcal L_{\mathrm{task}}$ provides the reference gradient $g_{\mathrm{task}}$ for evaluating segment-level OPD updates.

For each token $z_t$ in a completed response $z$, we compute the fixed advantage $A_t^{\mathrm{OPD}}$ using the log-probability difference in Eq.~\eqref{eq:opd_advantage}, with $\pi_T$ and $\pi_{\bar\theta}$ conditioned on $h_t^z$.
We use the following token-level OPD surrogate loss for all response tokens, including teacher guidance and Digest tokens:
\begin{equation}
\ell_t^{\mathrm{OPD}}(\theta)
=
-A_t^{\mathrm{OPD}}
\log\pi_\theta(z_t\mid h_t^z).
\label{eq:opd_loss}
\end{equation}
Let $\mathcal I$ be the set of response-token positions in the OPD training batch, excluding problem and padding positions.
Let $\mathcal S$ be the index set of all Digest segments in this batch, including the copied guidance segments.
For $i\in\mathcal S$, $\mathcal I_i\subseteq\mathcal I$ is the set of positions occupied by $D_i$, and $m_i=|\mathcal I_i|>0$ is its token count.
The segment gradient $g_i$ is computed from the mean OPD loss over the tokens in $D_i$:
\begin{equation}
g_i
=
\frac{1}{m_i}
\sum_{u\in\mathcal I_i}
\left.
\nabla_\theta\ell_u^{\mathrm{OPD}}(\theta)
\right|_{\theta=\bar\theta}.
\label{eq:segment_gradient}
\end{equation}
Here, $\ell_u^{\mathrm{OPD}}$ is the loss in Eq.~\eqref{eq:opd_loss} for the token at batch position $u$, evaluated at its own response prefix.
Both $g_i$ and $g_{\mathrm{task}}$ are taken with respect to all trainable student parameters.

Under a first-order approximation, a positive inner product $\langle g_i,g_{\mathrm{task}}\rangle$ predicts a decrease in $\mathcal L_{\mathrm{task}}$ after a small step along $-g_i$.
We use the nonnegative part of this inner product as the segment score~\cite{pmlr-v80-ren18a}:
\begin{equation}
s_i=\max\{0,\langle g_i,g_{\mathrm{task}}\rangle\}.
\label{eq:segment_score}
\end{equation}
The score depends on both the direction and norm of $g_i$.
When at least one segment score is positive, we normalize the scores to obtain segment weights (meta-weights) $w_i$, keeping the mean weight over all Digest-token positions equal to one:

\begin{equation}
w_i
=
\frac{s_i\sum_{j\in\mathcal S}m_j}
{\sum_{j\in\mathcal S}m_js_j}.
\label{eq:meta_weight}
\end{equation}
If all segment scores are zero, we set $w_i=1$ for every $i\in\mathcal S$.

\begin{table*}[!t]
\centering

\footnotesize
\setlength{\tabcolsep}{3.5pt}
\renewcommand{\arraystretch}{1.12}

\resizebox{\textwidth}{!}{%
\begin{tabular}{l*{9}{c}}
\toprule
\textbf{Method}
& \textbf{AIME2024}
& \textbf{AIME2025}
& \textbf{AIME2026}
& \textbf{MATH500}
& \textbf{AMC2023}
& \textbf{OlympiadBench}
& \textbf{HMMT-Feb26}
& \textbf{HMMT-Nov25}
& \textbf{Avg.} \\
\midrule

Teacher
& 60.42
& 46.04
& 52.19
& 94.20
& 93.83
& 70.62
& 31.25
& 41.88
& 61.30 \\

\midrule
\multicolumn{10}{c}{\textit{Student: Qwen3-0.6B-Non-Thinking}} \\
\midrule

Student
& 1.77
& 2.40
& 0.73
& 44.10
& 24.45
& 16.36
& 0.76
& 3.85
& 11.80 \\

SFT\textsubscript{@110}
& 4.90
& 7.60
& 4.06
& 59.45
& 34.92
& 26.74
& 1.89
& 3.02
& 17.82 \\

KD\textsubscript{@110}
& 4.17
& 7.19
& 4.79
& 57.75
& 35.23
& 27.15
& 2.94
& 2.71
& 17.74 \\

GRPO\textsubscript{@110}
& 8.23
& 15.21
& 10.00
& 68.60
& 46.56
& 34.42
& 7.86
& 6.04
& 24.62 \\

OPD\textsubscript{@110}
& 13.44
& 17.92
& 11.98
& 75.30
& 51.23
& 41.32
& 7.39
& 5.62
& 28.03 \\

TRD\textsubscript{@20}
& 4.58
& 8.54
& 3.96
& 57.60
& 36.33
& 26.93
& 3.79
& 2.81
& 18.07 \\

FastOPD\textsubscript{@100}
& 15.83
& \underline{20.10}
& 13.54
& 75.30
& 53.67
& 44.81
& \underline{11.55}
& 8.54
& 30.42 \\

SKD\textsubscript{@20}
& 8.44
& 16.98
& 9.79
& 66.65
& 43.67
& 35.76
& 8.14
& 5.62
& 24.38 \\

Relay-OPD\textsubscript{@75}
& \underline{15.94}
& \textbf{20.94}
& \underline{14.06}
& \underline{76.80}
& \underline{55.55}
& \underline{45.03}
& 11.17
& \underline{8.85}
& \underline{31.04} \\

\rowcolor{oursbg}
\textbf{CD-OPD}\textsubscript{@80}
& \textbf{16.77}
& \textbf{20.94}
& \textbf{15.63}
& \textbf{77.50}
& \textbf{56.33}
& \textbf{49.07}
& \textbf{12.97}
& \textbf{10.10}
& \textbf{32.41} \\

\addlinespace[2pt]
\rowcolor{improvementbg}
\textit{Improvement vs. OPD}
& $+3.33$
& $+3.02$
& $+3.65$
& $+2.20$
& $+5.10$
& $+7.75$
& $+5.58$
& $+4.48$
& $+4.38$ \\

\rowcolor{improvementbg}
\textit{Improvement vs. Relay-OPD}
& $+0.83$
& $0.00$
& $+1.57$
& $+0.70$
& $+0.78$
& $+4.04$
& $+1.80$
& $+1.25$
& $+1.37$ \\

\midrule
\multicolumn{10}{c}{\textit{Student: Qwen3-1.7B-Non-Thinking}} \\
\midrule

Student
& 12.60
& 9.58
& 7.40
& 71.95
& 47.89
& 38.54
& 6.34
& 4.38
& 24.84 \\

SFT\textsubscript{@110}
& 23.33
& 19.48
& 16.15
& 81.40
& 59.45
& 46.62
& 12.59
& 6.56
& 33.20 \\

KD\textsubscript{@110}
& 23.54
& 21.15
& 15.31
& 81.40
& 59.45
& 46.62
& 12.59
& 6.56
& 33.33 \\

GRPO\textsubscript{@110}
& 24.58
& 22.08
& 15.31
& 81.45
& 60.23
& 48.07
& 12.78
& 7.50
& 34.00 \\

OPD\textsubscript{@55}
& 35.83
& 25.52
& 23.33
& 85.70
& 70.08
& 55.27
& 20.08
& 14.06
& 41.23 \\

TRD\textsubscript{@40}
& 19.27
& 19.69
& 12.71
& 77.70
& 55.47
& 44.18
& 11.93
& 4.58
& 30.69 \\

FastOPD\textsubscript{@45}
& 42.29
& 30.42
& 26.35
& 87.95
& 74.30
& 58.16
& 23.58
& \textbf{20.73}
& 45.47 \\

SKD\textsubscript{@35}
& 33.12
& 30.73
& 28.85
& 87.35
& 72.42
& 54.41
& 20.08
& 11.88
& 42.35 \\

Relay-OPD\textsubscript{@35}
& \underline{42.71}
& \textbf{32.81}
& \underline{30.52}
& \underline{89.50}
& \underline{76.88}
& \underline{58.79}
& \underline{24.72}
& 19.79
& \underline{46.96} \\

\rowcolor{oursbg}
\textbf{CD-OPD}\textsubscript{@40}
& \textbf{43.65}
& \underline{31.56}
& \textbf{33.23}
& \textbf{89.80}
& \textbf{79.53}
& \textbf{63.46}
& \textbf{24.81}
& \underline{20.21}
& \textbf{48.28} \\

\addlinespace[2pt]
\rowcolor{improvementbg}
\textit{Improvement vs. OPD}
& $+7.82$
& $+6.04$
& $+9.90$
& $+4.10$
& $+9.45$
& $+8.19$
& $+4.73$
& $+6.15$
& $+7.05$ \\

\rowcolor{improvementbg}
\textit{Improvement vs. Relay-OPD}
& $+0.94$
& $-1.25$
& $+2.71$
& $+0.30$
& $+2.65$
& $+4.67$
& $+0.09$
& $+0.42$
& $+1.32$ \\

\bottomrule
\end{tabular}%
}
\caption{
Performance across eight mathematical reasoning benchmarks.
The best and second-best results within each student size are shown in bold and underlined, respectively.
Blue-shaded rows denote our method, and light-blue-shaded rows denote improvements in percentage points relative to the indicated baselines.
}
\label{main_results}
\end{table*}

\paragraph{Policy Update.}
For $u\in\mathcal I$, we set $\omega_u=w_i$ when $u\in\mathcal I_i$ and $\omega_u=1$ otherwise.
Weights above one strengthen a segment's OPD contribution, whereas weights below one weaken it.
The final objective is
\begin{equation}
\mathcal L_{\mathrm{CD\text{-}OPD}}(\theta)
=
\frac{1}{|\mathcal I|}
\sum_{u\in\mathcal I}
\omega_u\ell_u^{\mathrm{OPD}}(\theta).
\label{eq:weighted_opd}
\end{equation}
These segment weights emphasize Digest segments whose OPD updates are predicted to reduce $\mathcal L_{\mathrm{task}}$ more.
With the sampled responses, $A_{b,k}^{\mathrm{task}}$, $A_t^{\mathrm{OPD}}$, and $w_i$ treated as constants, we perform one optimizer step using the gradient of $\mathcal L_{\mathrm{CD\text{-}OPD}}$ at $\theta=\bar\theta$.
The next iteration uses the updated student policy.

\section{Experiments}
\paragraph{Models and Training Data.}
Following prior OPD studies~\citep{fastopd,relay-opd}, we adopt a mathematical reasoning setup with Qwen3-0.6B and Qwen3-1.7B as students and Qwen3-4B-Instruct-2507 as the teacher~\citep{yang2025qwen3}.
Both students operate in non-thinking mode, and the teacher remains frozen throughout training.
Training uses the English subset of DAPO-Math-17K~\citep{yu2025dapo}, which pairs mathematical reasoning problems with integer-valued reference answers. Dataset details are provided in Appendix~\ref{training_data}

\paragraph{Benchmarks and Metrics.}
We evaluate all methods on eight mathematical reasoning benchmarks:
AIME2024~\citep{aimo2024aime},
AIME2025~\citep{mathai2025aime},
AIME2026~\citep{mathai2026aime},
MATH500~\citep{lightman2024lets},
AMC2023~\citep{aimo2024amc},
OlympiadBench~\citep{he-etal-2024-olympiadbench},
HMMT-Feb26~\citep{dekoninck2026matharena},
and HMMT-Nov25~\citep{dekoninck2026matharena}.
We report mean@4 for MATH500 and OlympiadBench and mean@32 for the remaining benchmarks.
Avg. is the unweighted mean across benchmarks.
Dataset details are provided in Appendix~\ref{benchmark}

\paragraph{Implementation Details.} See Appendix~\ref{app:training_hyperparameters}.

\paragraph{Baselines.}
For reference, Teacher (frozen) and Student (before training) denote the released teacher and student checkpoints evaluated without further training.
We compare CD-OPD with supervised fine-tuning (SFT), knowledge distillation (KD)~\citep{KD}, and Group Relative Policy Optimization (GRPO)~\citep{grpo}.
We also include OPD~\citep{GDK}, TRD~\citep{trd}, FastOPD~\citep{fastopd}, SKD~\citep{SKD}, and Relay-OPD~\citep{relay-opd} as distillation baselines. @step denotes the reported checkpoint step.

\paragraph{Main Experimental Results.}
Table~\ref{main_results} shows that CD-OPD achieves the highest average accuracy among the compared training methods, reaching 32.41\% for Qwen3-0.6B and 48.28\% for Qwen3-1.7B.
The gains over standard OPD range from 4.38 to 7.05 percentage points and remain 1.32--1.37 points over Relay-OPD, the strongest baseline on average.
\textit{Both CD-OPD and Relay-OPD use the same reflection-triggered takeover criterion during training; the gains over Relay-OPD highlight the value of improving how students learn from teacher guidance and support CD-OPD's effectiveness in strengthening their independent reasoning.}
The advantage extends across benchmarks: CD-OPD achieves the best accuracy on all eight benchmarks with Qwen3-0.6B and on six with Qwen3-1.7B.
These results suggest that CD-OPD helps students learn more effectively from teacher guidance by digesting its earlier context and reweighting Digest-segment OPD losses to improve student-only performance.

\begin{table*}[!t]
\centering

\footnotesize
\setlength{\tabcolsep}{2pt}
\renewcommand{\arraystretch}{1.12}

\resizebox{\textwidth}{!}{%
\begin{tabular}{l*{10}{c}}
\toprule
\textbf{Method}
& \textbf{AIME2024}
& \textbf{AIME2025}
& \textbf{AIME2026}
& \textbf{MATH500}
& \textbf{AMC2023}
& \textbf{OlympiadBench}
& \textbf{HMMT-Feb26}
& \textbf{HMMT-Nov25}
& \textbf{Avg.}
& $\boldsymbol{\Delta}$ \\
\midrule

\multicolumn{11}{c}{
\textit{Student: Qwen3-0.6B-Non-Thinking}
} \\
\midrule

\rowcolor{oursbg}
\textbf{CD-OPD}
& \textbf{16.77}
& \textbf{20.94}
& \textbf{15.63}
& \textbf{77.50}
& \textbf{56.33}
& \textbf{49.07}
& \textbf{12.97}
& \textbf{10.10}
& \textbf{32.41}
& --- \\

\addlinespace[2pt]
w/o Dependency Graph
& 14.58
& \underline{18.33}
& 14.58
& 76.35
& 55.47
& \underline{48.74}
& \underline{11.65}
& 7.60
& 30.91
& \cellcolor{improvementbg}$-1.50$ \\

\addlinespace[2pt]
w/o Online Meta-Reweighting
& \underline{16.35}
& 17.50
& \underline{15.21}
& \underline{76.65}
& \underline{56.02}
& 48.70
& 11.46
& \underline{8.65}
& \underline{31.32}
& \cellcolor{improvementbg}$-1.09$ \\

\midrule
\multicolumn{11}{c}{
\textit{Student: Qwen3-1.7B-Non-Thinking}
} \\
\midrule

\rowcolor{oursbg}
\textbf{CD-OPD}
& \textbf{43.65}
& \textbf{31.56}
& \textbf{33.23}
& \textbf{89.80}
& \textbf{79.53}
& \textbf{63.46}
& \textbf{24.81}
& \textbf{20.21}
& \textbf{48.28}
& --- \\

\addlinespace[2pt]
w/o Dependency Graph
& \underline{43.23}
& \underline{31.25}
& \underline{30.73}
& \underline{88.75}
& \underline{78.26}
& \underline{62.05}
& \underline{24.43}
& \underline{20.00}
& \underline{47.34}
& \cellcolor{improvementbg}$-0.94$ \\

\addlinespace[2pt]
w/o Online Meta-Reweighting
& 41.98
& 30.73
& 30.00
& 88.05
& 78.06
& 61.28
& 24.04
& 19.56
& 46.71
& \cellcolor{improvementbg}$-1.57$ \\

\bottomrule
\end{tabular}%
}
\caption{
Ablation results across eight benchmarks.
$\Delta$ denotes the change in average accuracy relative to CD-OPD, in percentage points.
}
\label{tab:ablation}

\end{table*}

\paragraph{Ablation Study.}
Removing the Contextual Digest and its segment-specific OPD loss weighting reduces CD-OPD to the Relay-OPD formulation evaluated in Table~\ref{main_results}.
Table~\ref{tab:ablation} therefore examines dependency-graph-based information selection and online meta-reweighting separately.
The \textit{w/o Dependency Graph} variant directly summarizes the context without a dependency graph, while retaining online meta-reweighting.
Average accuracy decreases by 0.94--1.50 percentage points for different students, supporting the use of explicit dependencies to select context relevant to the current teacher guidance.
The \textit{w/o Online Meta-Reweighting} variant retains Contextual Digest construction but assigns unit weights to all Digest segments, reducing average accuracy by 1.09--1.57 points for different students.
Both settings keep the mean weight over Digest-token positions equal to one, so the gains support using the task gradient and segment gradients to adjust relative segment weights.
\textit{These results show that graph-based information selection and OPD loss weighting both contribute to CD-OPD's accuracy gains.}

\begin{figure}[t]
    \centering
    \includegraphics[width=\linewidth]{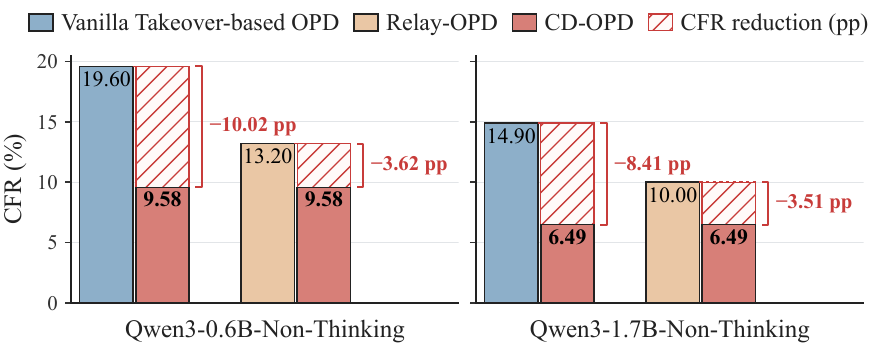}
\caption{
CFR for Qwen3-0.6B and Qwen3-1.7B, computed by
pooling numerator and denominator counts across eight benchmarks.
}
\label{Figure4}
\end{figure}

\paragraph{CFR Analysis.}
Figure~\ref{Figure4} compares CFR across eight mathematical reasoning benchmarks for Vanilla Takeover-based OPD, Relay-OPD, and CD-OPD.
CD-OPD achieves CFRs of 9.58\% and 6.49\% for Qwen3-0.6B and Qwen3-1.7B, respectively.
Compared with Vanilla Takeover-based OPD, CFR decreases by 10.02 and 8.41 percentage points, corresponding to relative reductions of 51.1\% and 56.4\%.
Relative to Relay-OPD, CD-OPD further reduces CFR by 3.62 and 3.51 percentage points, or 27.4\% and 35.1\%, respectively.
\textit{These reductions demonstrate that CD-OPD's combination of Contextual Digest construction and online meta-reweighting improves independent reasoning while mitigating collaboration failures under teacher takeover.}

\begin{figure}[t]
    \centering
    \includegraphics[width=\linewidth]{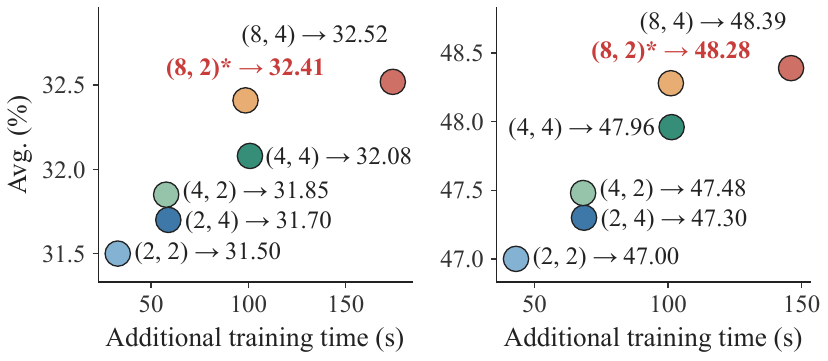}
\caption{
Effects of $B$ and $R$ on CD-OPD's average accuracy across eight benchmarks and additional training time relative to training without online meta-reweighting.
\textbf{Left}: Qwen3-0.6B; \textbf{Right}: Qwen3-1.7B.
}
\label{Figure5}
\end{figure}

\paragraph{Sensitivity to $B$ and $R$.}
Figure~\ref{Figure5} reports CD-OPD's average accuracy across eight benchmarks and additional training time spent on online meta-reweighting for $B\in\{2,4,8\}$ and $R\in\{2,4\}$.
With the same number of student-only responses, $(B,R)=(8,2)$ achieves 0.32--0.33 percentage points higher average accuracy than $(4,4)$ at comparable additional training time.
At $B=8$, increasing $R$ from 2 to 4 yields a further 0.11-point gain for both students, but increases additional training time by 44.5\%--76.8\%.
This cost also depends on student-only response length (at $(B,R)=(8,2)$, mean lengths are 8,974.95 and 5,330.61 tokens for Qwen3-0.6B and Qwen3-1.7B, respectively).
\textit{These comparisons suggest that distributing a fixed number of student-only responses across more training problems can improve the accuracy--time trade-off of CD-OPD's online meta-reweighting.}
Detailed results are provided in Appendix~\ref{app:br_results}.

\paragraph{Contextual Digest Length Analysis.}
\begin{figure}[t]
    \centering
    \includegraphics[width=0.85\linewidth]{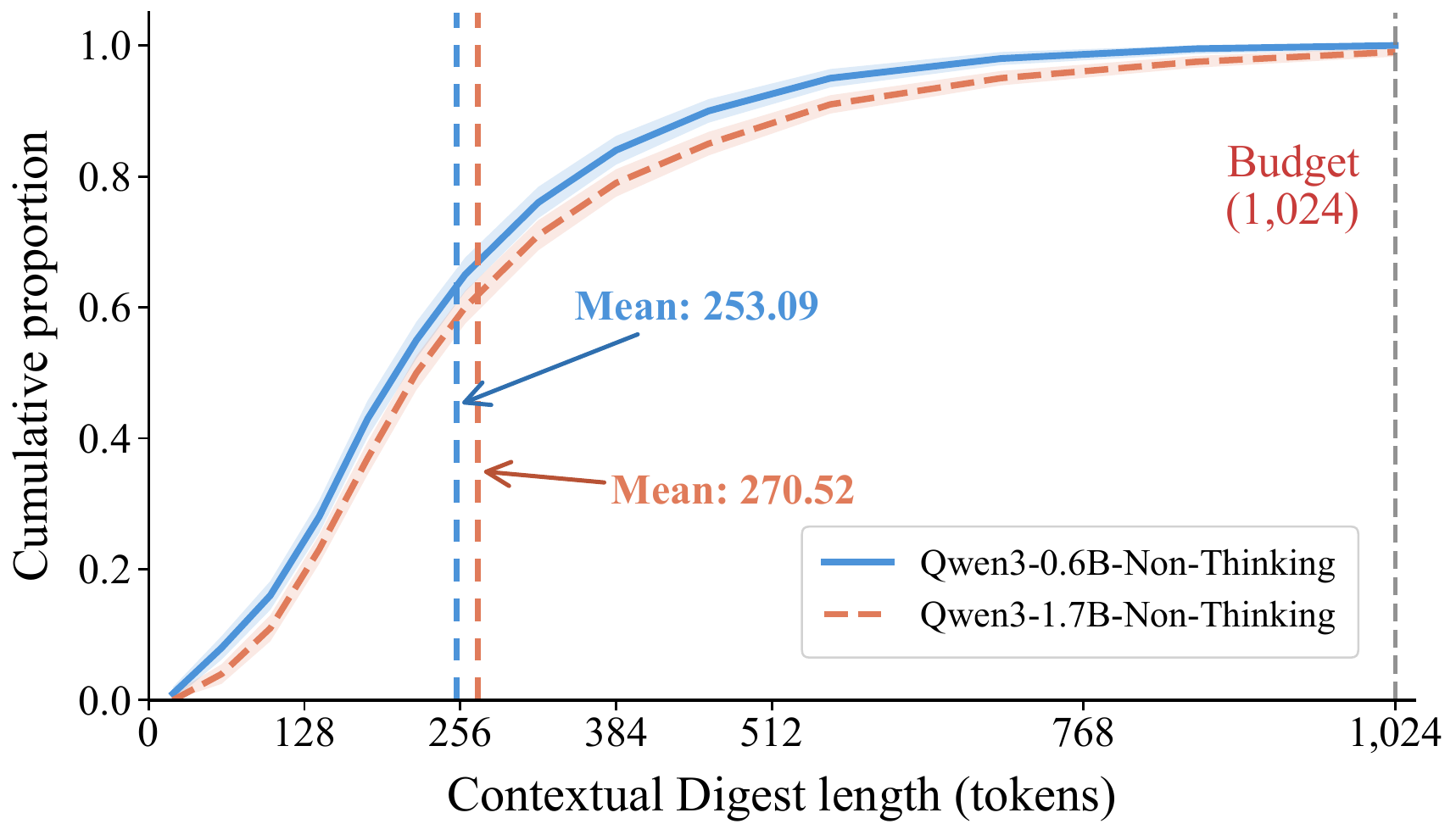}
    \caption{
Contextual Digest length distributions during training for Qwen3-0.6B and Qwen3-1.7B.
}
\label{Figure6}
\end{figure}
Figure~\ref{Figure6} presents the cumulative distribution of Contextual Digest lengths during training.
Under the 1,024-token budget, the mean Digest lengths per takeover are 253.09 and 270.52 tokens for Qwen3-0.6B and Qwen3-1.7B, respectively, while the maximum observed lengths reach 968 and 1,020 tokens.
\textit{These results show that CD-OPD comfortably uses approximately one quarter of the token budget per takeover on average, while accommodating longer Contextual Digests when needed.}

\paragraph{Training and Inference Efficiency.}
\textit{Relative to Relay-OPD, CD-OPD achieves higher average accuracy and lower CFR with comparable mean student-only response lengths and inference time.}
Detailed results are provided in Appendix~\ref{app:efficiency}.

\paragraph{Case Study.}
We compare student-only responses from Relay-OPD and CD-OPD on an AIME2025 problem.
Both students derive the correct rational product $31/13$, but the Relay-OPD student later mistakes one factor, $31$, for the complete product.
The resulting error stems from incorrectly using an established intermediate result rather than failing to derive it.
In contrast, the CD-OPD student uses the complete product in the final calculation and answers correctly.
Details and prompt templates are provided in
Appendices~\ref{app:case_study} and~\ref{app:prompt_templates},
respectively.

\section{Conclusion}
Our CFR-based diagnosis reveals failures under teacher takeover that overall accuracy can obscure, and we propose CD-OPD to address them through trajectory construction and policy update in a novel way.
Contextual Digest uses a teacher-generated dependency graph to organize and summarize teacher guidance with its earlier context, while online meta-reweighting uses the task gradient and segment gradients to determine each Digest segment's OPD loss weight.
Experiments on eight benchmarks show higher average accuracy and lower CFR than the baselines.
Together, these findings highlight the value of addressing both how teacher guidance is curated for student continuation and how the resulting OPD supervision is adapted to improve students' independent reasoning, while maintaining comparable inference efficiency.

\section*{Limitations}
We identify two limitations of the current CD-OPD design.
First, Contextual Digest construction uses fixed limits on dependency-graph size and Digest length.
These limits may restrict the contextual information retained when teacher guidance depends on many problem conditions or intermediate results.
A possible extension is to adapt the construction budget to the amount of earlier context required by the current guidance, rather than applying the same limit at every takeover.
Second, online meta-reweighting assigns one meta-weight to each Digest segment based on its segment gradient and the task gradient.
All tokens within that segment therefore share the same meta-weight, although their individual OPD gradients may differ in alignment with the task gradient.
Finer-grained weighting could account for these differences, but its benefits for students' independent reasoning need to be evaluated against the additional gradient-computation cost.

\bibliography{custom}

\begin{thebibliography}{29}
\providecommand{\natexlab}[1]{#1}

\bibitem[{Agarwal et~al.(2024)Agarwal, Vieillard, Zhou, Stanczyk, Garea, Geist,
  and Bachem}]{GDK}
Rishabh Agarwal, Nino Vieillard, Yongchao Zhou, Piotr Stanczyk, Sabela~Ramos
  Garea, Matthieu Geist, and Olivier Bachem. 2024.
\newblock \href {https://openreview.net/forum?id=3zKtaqxLhW} {On-policy
  distillation of language models: Learning from self-generated mistakes}.
\newblock In \emph{The Twelfth International Conference on Learning
  Representations}.

\bibitem[{{AI-MO}(2024{\natexlab{a}})}]{aimo2024aime}
{AI-MO}. 2024{\natexlab{a}}.
\newblock \href {https://huggingface.co/datasets/AI-MO/aimo-validation-aime}
  {{AIMO Validation AIME Dataset}}.

\bibitem[{{AI-MO}(2024{\natexlab{b}})}]{aimo2024amc}
{AI-MO}. 2024{\natexlab{b}}.
\newblock \href {https://huggingface.co/datasets/AI-MO/aimo-validation-amc}
  {{AIMO Validation AMC Dataset}}.

\bibitem[{Dekoninck et~al.(2026)Dekoninck, Jovanovi{\'c}, Gehrunger,
  R{\"o}gnvaldsson, Petrov, Sun, and Vechev}]{dekoninck2026matharena}
Jasper Dekoninck, Nikola Jovanovi{\'c}, Tim Gehrunger, K{\'a}ri
  R{\"o}gnvaldsson, Ivo Petrov, Chenhao Sun, and Martin Vechev. 2026.
\newblock \href {https://arxiv.org/abs/2605.00674} {Beyond benchmarks:
  Matharena as an evaluation platform for mathematics with llms}.
\newblock \emph{Preprint}, arXiv:2605.00674.

\bibitem[{Fu et~al.(2026)Fu, Huang, Jiang, Liu, Jiang, Zhu, and
  Zhao}]{fu2026revisiting}
Yuqian Fu, Haohuan Huang, Kaiwen Jiang, Jiacai Liu, Zhuo Jiang, Yuanheng Zhu,
  and Dongbin Zhao. 2026.
\newblock \href {https://openreview.net/forum?id=uZbZb058bA} {Revisiting
  on-policy distillation: Empirical failure modes and simple fixes}.
\newblock In \emph{Third Conference on Language Modeling}.

\bibitem[{Gu et~al.(2026)Gu, Dong, Wei, and Huang}]{MiniLLM}
Yuxian Gu, Li~Dong, Furu Wei, and Minlie Huang. 2026.
\newblock \href {https://arxiv.org/abs/2306.08543} {Minillm: On-policy
  distillation of large language models}.
\newblock \emph{Preprint}, arXiv:2306.08543.

\bibitem[{He et~al.(2024)He, Luo, Bai, Hu, Thai, Shen, Hu, Han, Huang, Zhang,
  Liu, Qi, Liu, and Sun}]{he-etal-2024-olympiadbench}
Chaoqun He, Renjie Luo, Yuzhuo Bai, Shengding Hu, Zhen Thai, Junhao Shen, Jinyi
  Hu, Xu~Han, Yujie Huang, Yuxiang Zhang, Jie Liu, Lei Qi, Zhiyuan Liu, and
  Maosong Sun. 2024.
\newblock \href {https://doi.org/10.18653/v1/2024.acl-long.211}
  {{O}lympiad{B}ench: A challenging benchmark for promoting {AGI} with
  olympiad-level bilingual multimodal scientific problems}.
\newblock In \emph{Proceedings of the 62nd Annual Meeting of the Association
  for Computational Linguistics (Volume 1: Long Papers)}, pages 3828--3850,
  Bangkok, Thailand. Association for Computational Linguistics.

\bibitem[{Hinton et~al.(2015)Hinton, Vinyals, and Dean}]{KD}
Geoffrey Hinton, Oriol Vinyals, and Jeff Dean. 2015.
\newblock \href {https://arxiv.org/abs/1503.02531} {Distilling the knowledge in
  a neural network}.
\newblock \emph{Preprint}, arXiv:1503.02531.

\bibitem[{Hsieh et~al.(2023)Hsieh, Li, Yeh, Nakhost, Fujii, Ratner, Krishna,
  Lee, and Pfister}]{hsieh-etal-2023-distilling}
Cheng-Yu Hsieh, Chun-Liang Li, Chih-kuan Yeh, Hootan Nakhost, Yasuhisa Fujii,
  Alex Ratner, Ranjay Krishna, Chen-Yu Lee, and Tomas Pfister. 2023.
\newblock \href {https://doi.org/10.18653/v1/2023.findings-acl.507} {Distilling
  step-by-step! outperforming larger language models with less training data
  and smaller model sizes}.
\newblock In \emph{Findings of the Association for Computational Linguistics:
  ACL 2023}, pages 8003--8017, Toronto, Canada. Association for Computational
  Linguistics.

\bibitem[{Jiang et~al.(2026)Jiang, Xu, Ding, and Zhang}]{trd}
Li~Jiang, Haoran Xu, Yichuan Ding, and Amy Zhang. 2026.
\newblock \href {https://arxiv.org/abs/2606.08432} {Trajectory-refined
  distillation}.
\newblock \emph{Preprint}, arXiv:2606.08432.

\bibitem[{Kim and Rush(2016)}]{SequenceKD}
Yoon Kim and Alexander~M. Rush. 2016.
\newblock \href {https://doi.org/10.18653/v1/D16-1139} {Sequence-level
  knowledge distillation}.
\newblock In \emph{Proceedings of the 2016 Conference on Empirical Methods in
  Natural Language Processing}, pages 1317--1327, Austin, Texas. Association
  for Computational Linguistics.

\bibitem[{Koo et~al.(2025)Koo, Hwang, Kim, Kang, Bae, and Jung}]{switch}
Jahyun Koo, Yerin Hwang, Yongil Kim, Taegwan Kang, Hyunkyung Bae, and Kyomin
  Jung. 2025.
\newblock \href {https://doi.org/10.18653/v1/2025.findings-naacl.206}
  {{SWITCH}: Studying with teacher for knowledge distillation of large language
  models}.
\newblock In \emph{Findings of the Association for Computational Linguistics:
  NAACL 2025}, pages 3733--3746, Albuquerque, New Mexico. Association for
  Computational Linguistics.

\bibitem[{Kullback and Leibler(1951)}]{kullback1951information}
Solomon Kullback and Richard~A Leibler. 1951.
\newblock On information and sufficiency.
\newblock \emph{The annals of mathematical statistics}, 22(1):79--86.

\bibitem[{Li et~al.(2026)Li, Zuo, He, Zhang, Xiao, Qian, Yu, ang Gao, Yang,
  Liu, and Ding}]{rethinking_opd}
Yaxuan Li, Yuxin Zuo, Bingxiang He, Jinqian Zhang, Chaojun Xiao, Cheng Qian,
  Tianyu Yu, Huan ang Gao, Wenkai Yang, Zhiyuan Liu, and Ning Ding. 2026.
\newblock \href {https://arxiv.org/abs/2604.13016} {Rethinking on-policy
  distillation of large language models: Phenomenology, mechanism, and recipe}.
\newblock \emph{Preprint}, arXiv:2604.13016.

\bibitem[{Lightman et~al.(2023)Lightman, Kosaraju, Burda, Edwards, Baker, Lee,
  Leike, Schulman, Sutskever, and Cobbe}]{lightman2024lets}
Hunter Lightman, Vineet Kosaraju, Yura Burda, Harri Edwards, Bowen Baker, Teddy
  Lee, Jan Leike, John Schulman, Ilya Sutskever, and Karl Cobbe. 2023.
\newblock \href {https://arxiv.org/abs/2305.20050} {Let's verify step by step}.
\newblock \emph{Preprint}, arXiv:2305.20050.

\bibitem[{Ren et~al.(2018)Ren, Zeng, Yang, and Urtasun}]{pmlr-v80-ren18a}
Mengye Ren, Wenyuan Zeng, Bin Yang, and Raquel Urtasun. 2018.
\newblock \href {https://proceedings.mlr.press/v80/ren18a.html} {Learning to
  reweight examples for robust deep learning}.
\newblock In \emph{Proceedings of the 35th International Conference on Machine
  Learning}, volume~80 of \emph{Proceedings of Machine Learning Research},
  pages 4334--4343. PMLR.

\bibitem[{Shao et~al.(2024)Shao, Wang, Zhu, Xu, Song, Bi, Zhang, Zhang, Li, Wu,
  and Guo}]{grpo}
Zhihong Shao, Peiyi Wang, Qihao Zhu, Runxin Xu, Junxiao Song, Xiao Bi, Haowei
  Zhang, Mingchuan Zhang, Y.~K. Li, Y.~Wu, and Daya Guo. 2024.
\newblock \href {https://arxiv.org/abs/2402.03300} {Deepseekmath: Pushing the
  limits of mathematical reasoning in open language models}.
\newblock \emph{Preprint}, arXiv:2402.03300.

\bibitem[{Song and Zheng(2026)}]{opd1}
Mingyang Song and Mao Zheng. 2026.
\newblock \href {https://arxiv.org/abs/2604.00626} {A survey of on-policy
  distillation for large language models}.
\newblock \emph{Preprint}, arXiv:2604.00626.

\bibitem[{Wang et~al.(2026)Wang, Yan, Shen, kaiyuan liu, Fan, Li, Miao, Yuan,
  Shen, and Ye}]{motab}
Bing Wang, Shaotian Yan, Chen Shen, kaiyuan liu, Sinan Fan, Ximing Li, Rui
  Miao, Xiaosong Yuan, Zhanming Shen, and Jieping Ye. 2026.
\newblock \href {https://arxiv.org/abs/2605.19433} {Backtracking when it
  strays: Mitigating dual exposure biases in llm reasoning distillation}.
\newblock \emph{Preprint}, arXiv:2605.19433.

\bibitem[{Xu et~al.(2026)Xu, Xu, Hong, Ni, Li, Qiu, Lu, and Shen}]{relay-opd}
Haolei Xu, Xiaowen Xu, Haiwen Hong, Zixuan Ni, Hongxing Li, Yiwen Qiu, Weiming
  Lu, and Yongliang Shen. 2026.
\newblock \href {https://arxiv.org/abs/2607.26057} {Pass the baton:
  Trajectory-relayed on-policy distillation}.
\newblock \emph{Preprint}, arXiv:2607.26057.

\bibitem[{Xu et~al.(2025)Xu, Han, Wang, Le, Madeka, Li, Wang, Agarwal, Lee, and
  Pfister}]{SKD}
Wenda Xu, Rujun Han, Zifeng Wang, Long Le, Dhruv Madeka, Lei Li, William~Yang
  Wang, Rishabh Agarwal, Chen-Yu Lee, and Tomas Pfister. 2025.
\newblock \href {https://openreview.net/forum?id=EgJhwYR2tB} {Speculative
  knowledge distillation: Bridging the teacher-student gap through interleaved
  sampling}.
\newblock In \emph{The Thirteenth International Conference on Learning
  Representations}.

\bibitem[{Yang et~al.(2025)Yang, Li, Yang, Zhang, Hui, Zheng, Yu, Gao, Huang,
  Lv, Zheng, Liu, Zhou, Huang, Hu, Ge, Wei, Lin, Tang, Yang, Tu, Zhang, Yang,
  Yang, Zhou, Zhou, Lin, Dang, Bao, Yang, Yu, Deng, Li, Xue, Li, Zhang, Wang,
  Zhu, Men, Gao, Liu, Luo, Li, Tang, Yin, Ren, Wang, Zhang, Ren, Fan, Su,
  Zhang, Zhang, Wan, Liu, Wang, Cui, Zhang, Zhou, and Qiu}]{yang2025qwen3}
An~Yang, Anfeng Li, Baosong Yang, Beichen Zhang, Binyuan Hui, Bo~Zheng, Bowen
  Yu, Chang Gao, Chengen Huang, Chenxu Lv, Chujie Zheng, Dayiheng Liu, Fan
  Zhou, Fei Huang, Feng Hu, Hao Ge, Haoran Wei, Huan Lin, Jialong Tang, and 41
  others. 2025.
\newblock \href {https://arxiv.org/abs/2505.09388} {Qwen3 technical report}.
\newblock \emph{Preprint}, arXiv:2505.09388.

\bibitem[{Yu et~al.(2026)Yu, Xu, Xu, Zhou, and Lin}]{opd2}
Fangxu Yu, Weijia Xu, Michael Xu, Tianyi Zhou, and Zinan Lin. 2026.
\newblock \href {https://arxiv.org/abs/2607.26246} {Weak-to-strong on-policy
  distillation}.
\newblock \emph{Preprint}, arXiv:2607.26246.

\bibitem[{Yu et~al.(2025)Yu, Zhang, Zhu, Yuan, Zuo, Yue, Dai, Fan, Liu, Liu,
  Liu, Lin, Lin, Ma, Sheng, Tong, Zhang, Zhang, Zhang, Zhu, Zhu, Chen, Chen,
  Wang, Yu, Song, Wei, Zhou, Liu, Ma, Zhang, Yan, Qiao, Wu, and
  Wang}]{yu2025dapo}
Qiying Yu, Zheng Zhang, Ruofei Zhu, Yufeng Yuan, Xiaochen Zuo, Yu~Yue, Weinan
  Dai, Tiantian Fan, Gaohong Liu, Lingjun Liu, Xin Liu, Haibin Lin, Zhiqi Lin,
  Bole Ma, Guangming Sheng, Yuxuan Tong, Chi Zhang, Mofan Zhang, Wang Zhang,
  and 16 others. 2025.
\newblock \href {https://arxiv.org/abs/2503.14476} {Dapo: An open-source llm
  reinforcement learning system at scale}.
\newblock \emph{Preprint}, arXiv:2503.14476.

\bibitem[{Zhang et~al.(2026)Zhang, Yang, Janghorbani, Han, II, Qian, Lyng,
  Batra, and Tillman}]{fastopd}
Dongxu Zhang, Zhichao Yang, Sepehr Janghorbani, Jun Han, Andrew~Ressler II,
  Qian Qian, Gregory~D Lyng, Sanjit~Singh Batra, and Robert~E. Tillman. 2026.
\newblock \href {https://doi.org/10.18653/v1/2026.findings-acl.1276} {Fast and
  effective on-policy distillation from reasoning prefixes}.
\newblock In \emph{Findings of the {A}ssociation for {C}omputational
  {L}inguistics: {ACL} 2026}, pages 25553--25569, San Diego, California, United
  States. Association for Computational Linguistics.

\bibitem[{Zhang and {Math-AI Team}(2025)}]{mathai2025aime}
Yifan Zhang and {Math-AI Team}. 2025.
\newblock \href {https://huggingface.co/datasets/math-ai/aime25} {American
  invitational mathematics examination ({AIME}) 2025}.

\bibitem[{Zhang and {Math-AI Team}(2026)}]{mathai2026aime}
Yifan Zhang and {Math-AI Team}. 2026.
\newblock \href {https://huggingface.co/datasets/math-ai/aime26} {American
  invitational mathematics examination ({AIME}) 2026}.

\bibitem[{Zhao et~al.(2026)Zhao, Chen, Lin, Winata, Yao, and Tang}]{opd3}
Hanyang Zhao, Haoxian Chen, Han Lin, Genta~Indra Winata, David Yao, and Wenpin
  Tang. 2026.
\newblock \href {https://arxiv.org/abs/2606.01039} {Opd+: Rethinking the
  advantage design for on-policy distillation}.
\newblock \emph{Preprint}, arXiv:2606.01039.

\bibitem[{Ziheng et~al.(2026)Ziheng, Li, Tang, Wu, and Terzopoulos}]{opd_esr}
Zhou Ziheng, Jiaqi Li, Huacong Tang, Ying~Nian Wu, and Demetri Terzopoulos.
  2026.
\newblock \href {https://arxiv.org/abs/2605.27028} {Less is more: Early
  stopping rollout for on-policy distillation}.
\newblock \emph{Preprint}, arXiv:2605.27028.

\end{thebibliography}

\clearpage
\appendix
\section{Dataset Details}
\label{app:datasets}

\subsection{Training Dataset}
\label{training_data}

We use the English subset of DAPO-Math-17K~\citep{yu2025dapo} for training.
This subset contains 14,116 mathematical reasoning problems
paired with integer-valued reference answers.
Problem statements are used for trajectory construction, while reference answers are used to assess final-answer correctness.
At each iteration, CD-OPD samples problems for trajectory construction and online meta-reweighting from the same training subset, without overlap between the two batches. 
The original DAPO-Math-17K release declares an Apache-2.0 license.\footnote{\url{https://huggingface.co/datasets/BytedTsinghua-SIA/DAPO-Math-17k}}
The processed English subset does not specify a separate license.\footnote{\url{https://huggingface.co/datasets/open-r1/DAPO-Math-17k-Processed}}

\subsection{Evaluation Benchmarks}
\label{benchmark}

We evaluate on eight mathematical reasoning benchmarks. 
Responses are sampled with a temperature of 1.0,
top-$p$ of 1.0, and a maximum response length of 32,768 tokens.
Table~\ref{tab:dataset_statistics} summarizes the dataset versions used in our experiments, including their sizes, the number of responses sampled per problem, and their declared licenses.

\paragraph{AIME2024, AIME2025, and AIME2026.}
Each benchmark contains 30 problems from the American Invitational Mathematics Examination in the corresponding year.
We use the 2024 subset of the AI-MO AIME collection~\citep{aimo2024aime} and the MathArena releases for AIME2025 and AIME2026~\citep{dekoninck2026matharena}.

\paragraph{MATH500.}
MATH500~\citep{lightman2024lets} is a 500-problem evaluation subset of the MATH benchmark, covering seven mathematical subjects and five difficulty levels.
We use the complete MATH500 test split released by HuggingFaceH4.

\paragraph{AMC2023.}
We use the 2023 subset of the AI-MO AMC collection~\citep{aimo2024amc}, comprising 40 American Mathematics Competitions problems adapted to an integer-answer format.

\paragraph{OlympiadBench.}
OlympiadBench~\citep{he-etal-2024-olympiadbench} contains mathematics and physics problems across multiple languages and modalities.
We use its English, text-only, open-ended mathematics subset (\texttt{OE\_TO\_maths\_en\_COMP}), which contains 674 problems.

\paragraph{HMMT-Feb26 and HMMT-Nov25.}
We use the MathArena releases~\citep{dekoninck2026matharena} of problems from the February 2026 and November 2025 Harvard--MIT Mathematics Tournaments.
HMMT-Feb26 contains 33 problems, and HMMT-Nov25 contains 30.

\begin{table}[t]
\centering
\small
\setlength{\tabcolsep}{3pt}
\renewcommand{\arraystretch}{1.1}
\begin{tabular}{@{}lrrl@{}}
\toprule
\textbf{Benchmark} & $\boldsymbol{N}$ & $\boldsymbol{K}$ & \textbf{License} \\
\midrule
\href{https://huggingface.co/datasets/AI-MO/aimo-validation-aime}{AIME2024}
& 30 & 32 & Apache-2.0 \\
\href{https://huggingface.co/datasets/MathArena/aime_2025}{AIME2025}
& 30 & 32 & CC BY-NC-SA 4.0 \\
\href{https://huggingface.co/datasets/MathArena/aime_2026}{AIME2026}
& 30 & 32 & CC BY-NC-SA 4.0 \\
\href{https://huggingface.co/datasets/HuggingFaceH4/MATH-500}{MATH500}
& 500 & 4 & MIT \\
\href{https://huggingface.co/datasets/AI-MO/aimo-validation-amc}{AMC2023}
& 40 & 32 & Apache-2.0 \\
\href{https://huggingface.co/datasets/Hothan/OlympiadBench}{OlympiadBench}
& 674 & 4 & Apache-2.0 \\
\href{https://huggingface.co/datasets/MathArena/hmmt_feb_2026}{HMMT-Feb26}
& 33 & 32 & CC BY-NC-SA 4.0 \\
\href{https://huggingface.co/datasets/MathArena/hmmt_nov_2025}{HMMT-Nov25}
& 30 & 32 & CC BY-NC-SA 4.0 \\
\bottomrule
\end{tabular}
\caption{
Evaluation dataset statistics and declared licenses.
}
\label{tab:dataset_statistics}
\end{table}
\begin{center}
\small
\setlength{\tabcolsep}{3pt}
\renewcommand{\arraystretch}{1.1}
\begin{tabular}{@{}p{0.64\columnwidth}p{0.30\columnwidth}@{}}
\toprule
\textbf{Hyperparameter} & \textbf{Value} \\
\midrule
Student generation mode & Non-thinking \\
Maximum prompt length & 2,048 tokens \\
Maximum training response length & 16,384 tokens \\
Sampling temperature & 1.0 \\
Sampling top-$p$ & 1.0 \\
Responses per problem for trajectory construction & 1 \\
Global trajectory-construction batch size & 128 \\
PPO mini-batch size & 128 \\
PPO epochs per batch & 1 \\
PPO clipping parameter $\epsilon$ & 0.2 \\
Optimizer & AdamW \\
Learning rate & $1\times10^{-6}$ \\
Weight decay & 0.01 \\
AdamW $(\beta_1,\beta_2)$ & $(0.9,0.999)$ \\
Maximum gradient norm & 1.0 \\
Student optimizer steps per iteration & 1 \\
Teacher optimizer steps & 0 \\
Training epoch budget & 1 \\
Training seed & 42 \\
Checkpoint saving interval & 5 steps \\
\bottomrule
\end{tabular}
\captionof{table}{Training and optimization hyperparameters for CD-OPD.}
\label{tab:common_training_hyperparameters}
\end{center}

\section{Implementation Details}
\label{app:training_hyperparameters}

\subsection{Training Settings}
\label{app:training_settings}

Table~\ref{tab:common_training_hyperparameters} summarizes the training and optimization hyperparameters.
We optimize the student with AdamW.
Following Relay-OPD~\citep{relay-opd}, we use a training seed of 42
and report results from the best checkpoint for each student size.
Each iteration uses a global trajectory-construction batch of
128 problems, with one response generated per problem.
Student generation and teacher guidance use the same sampling
temperature and top-$p$.
All CD-OPD experiments are conducted on four NVIDIA RTX PRO 6000 GPUs
with 96\,GB of memory each and 88 Intel Xeon Platinum 8470Q vCPUs,
using Ubuntu 22.04, Python 3.10, PyTorch 2.1.2, and CUDA 11.8.
Following Relay-OPD~\citep{relay-opd}, we evaluate student-only
accuracy every five training steps and select the checkpoint with
the highest average accuracy across the eight benchmarks.
The selected checkpoints are at steps 80 and 40 for Qwen3-0.6B
and Qwen3-1.7B.

\paragraph{Policy-loss implementation.}
We retain the PPO-style clipped token-level surrogate used by
Relay-OPD~\citep{relay-opd}.
For each response token, the probability ratio is
\begin{equation}
\rho_t(\theta)
=
\frac{\pi_\theta(z_t\mid h_t^z)}
     {\pi_{\bar\theta}(z_t\mid h_t^z)}.
\label{eq:app_policy_ratio}
\end{equation}
The denominator is evaluated under the fixed student snapshot
on the same prefix, including for teacher guidance and Digest tokens.
The implemented token-level loss is
\begin{equation}
\begin{aligned}
\ell_t^{\mathrm{clip}}(\theta)
&= -\min\Bigl\{
\rho_t(\theta)A_t^{\mathrm{OPD}},\\
&\quad
\operatorname{clip}
\bigl(\rho_t(\theta),1-\epsilon,1+\epsilon\bigr)
A_t^{\mathrm{OPD}}
\Bigr\}.
\end{aligned}
\label{eq:app_clipped_surrogate}
\end{equation}

The PPO mini-batch contains the full global trajectory-construction
batch, with one optimization pass per batch.
Task and segment gradients are computed before any parameter update.
After computing the meta-weights, we accumulate the final weighted
OPD gradient and perform one student optimizer step.
The gradients are therefore evaluated at $\theta=\bar\theta$,
where $\rho_t(\bar\theta)=1$ lies inside the clipping interval.
With fixed OPD advantages,
\begin{equation}
\left.
\nabla_\theta\ell_t^{\mathrm{clip}}(\theta)
\right|_{\theta=\bar\theta}
=
\left.
\nabla_\theta\ell_t^{\mathrm{OPD}}(\theta)
\right|_{\theta=\bar\theta}.
\label{eq:app_surrogate_gradient}
\end{equation}
With the same token masks, loss normalization, and fixed meta-weights,
the two surrogates yield the same segment gradients and weighted
OPD gradient at this update point.
The final weighted OPD gradient is clipped to a maximum norm of 1.0
before the AdamW step.

\subsection{Trajectory Construction with Contextual Digest}

Table~\ref{tab:digest_hyperparameters} summarizes the teacher takeover and Contextual Digest construction hyperparameters.
Following prior OPD studies~\citep{relay-opd}, the reflection-token set
$\mathcal R$ is formed from
\textit{Wait}, \textit{But}, \textit{Hmm}, \textit{Actually},
\textit{Hold}, \textit{However}, \textit{Yet}, \textit{Oh},
\textit{Alternatively}, \textit{No}, \textit{Ah},
\textit{Oops}, and \textit{Well},
together with their capitalization and leading-space variants.
Teacher-generated segments are paragraphs delimited by blank lines.
The segment and token limits apply to each takeover.

\begin{center}
\small
\setlength{\tabcolsep}{3pt}
\renewcommand{\arraystretch}{1.1}
\begin{tabular}{@{}p{0.64\columnwidth}p{0.30\columnwidth}@{}}
\toprule
\textbf{Hyperparameter} & \textbf{Value} \\
\midrule
\multicolumn{2}{@{}l}{\textit{Teacher takeover}} \\
Student top-$K$ for takeover & $K=5$ \\
Maximum takeovers per response & $M=2$ \\
Maximum teacher-generated segments per takeover & $L=2$ \\
Maximum teacher guidance length per takeover & 256 tokens \\
\midrule
\multicolumn{2}{@{}l}{\textit{Contextual Digest construction}} \\
Dependency-graph generation temperature & 0 \\
Maximum dependency-graph generation length & 2,048 tokens \\
Maximum dependency-graph nodes & 32 \\
Maximum Contextual Digest length & 1,024 tokens \\
\bottomrule
\end{tabular}
\captionof{table}{Teacher takeover and Contextual Digest construction hyperparameters.}
\label{tab:digest_hyperparameters}
\end{center}

\subsection{Online Meta-Reweighting}

Table~\ref{tab:meta_hyperparameters} summarizes the online meta-reweighting hyperparameters.
At each iteration, we sample $B=8$ distinct problems from the same
training subset, excluding those in the trajectory-construction batch,
and generate $R=2$ student-only responses per problem.
These responses use the same sampling temperature, top-$p$,
and maximum training response length as trajectory construction.
Final-answer correctness is assessed against the corresponding
reference answers.

\begin{center}
\small
\setlength{\tabcolsep}{3pt}
\renewcommand{\arraystretch}{1.1}
\begin{tabular}{@{}p{0.64\columnwidth}p{0.30\columnwidth}@{}}
\toprule
\textbf{Hyperparameter} & \textbf{Value} \\
\midrule
Additional training problems per iteration & $B=8$ \\
Student-only responses per problem & $R=2$ \\
Maximum student-only response length & 16,384 tokens \\
\bottomrule
\end{tabular}
\captionof{table}{Online meta-reweighting hyperparameters.}
\label{tab:meta_hyperparameters}
\end{center}

\begin{table*}[t]
\centering

\footnotesize
\setlength{\tabcolsep}{3.5pt}
\renewcommand{\arraystretch}{1.12}

\resizebox{\textwidth}{!}{%
\begin{tabular}{c*{9}{c}}
\toprule
$\boldsymbol{(B,R)}$
& \textbf{AIME2024}
& \textbf{AIME2025}
& \textbf{AIME2026}
& \textbf{MATH500}
& \textbf{AMC2023}
& \textbf{OlympiadBench}
& \textbf{HMMT-Feb26}
& \textbf{HMMT-Nov25}
& \textbf{Avg.} \\
\midrule

\multicolumn{10}{c}{
\textit{Student: Qwen3-0.6B-Non-Thinking}
} \\
\midrule

$(2,2)$
& 16.42 & 18.07 & 15.28 & 76.80
& 56.07 & 48.76 & 11.71 & 8.89
& 31.50 \\

$(4,2)$
& 16.55 & 19.17 & 15.41 & 77.06
& 56.17 & 48.88 & 12.19 & 9.35
& 31.85 \\

\rowcolor{oursbg}
$\boldsymbol{(8,2)}$
& \textbf{16.77}
& \textbf{20.94}
& \textbf{15.63}
& \textbf{77.50}
& \textbf{56.33}
& \textbf{49.07}
& \textbf{12.97}
& \textbf{10.10}
& \textbf{32.41} \\

\addlinespace[2pt]
$(2,4)$
& 16.50 & 18.70 & 15.36 & 76.95
& 56.13 & 48.83 & 11.99 & 9.16
& 31.70 \\

$(4,4)$
& 16.64 & 19.89 & 15.50 & 77.24
& 56.24 & 48.96 & 12.51 & 9.66
& 32.08 \\

$(8,4)$
& 16.81 & 21.27 & 15.67 & 77.58
& 56.36 & 49.11 & 13.12 & 10.24
& 32.52 \\

\midrule
\multicolumn{10}{c}{
\textit{Student: Qwen3-1.7B-Non-Thinking}
} \\
\midrule

$(2,2)$
& 42.29 & 30.87 & 30.59 & 88.37
& 78.33 & 61.68 & 24.18 & 19.68
& 47.00 \\

$(4,2)$
& 42.80 & 31.11 & 31.59 & 88.91
& 78.78 & 62.35 & 24.42 & 19.88
& 47.48 \\

\rowcolor{oursbg}
$\boldsymbol{(8,2)}$
& \textbf{43.65}
& \textbf{31.56}
& \textbf{33.23}
& \textbf{89.80}
& \textbf{79.53}
& \textbf{63.46}
& \textbf{24.81}
& \textbf{20.21}
& \textbf{48.28} \\

\addlinespace[2pt]
$(2,4)$
& 42.61 & 31.02 & 31.22 & 88.71
& 78.61 & 62.10 & 24.33 & 19.80
& 47.30 \\

$(4,4)$
& 43.31 & 31.35 & 32.58 & 89.45
& 79.23 & 63.02 & 24.66 & 20.08
& 47.96 \\

$(8,4)$
& 43.77 & 31.56 & 33.47 & 89.93
& 79.64 & 63.62 & 24.87 & 20.26
& 48.39 \\

\bottomrule
\end{tabular}%
}
\caption{
Per-benchmark accuracy (\%) of CD-OPD for different $(B,R)$ configurations.
Bold rows indicate the default setting $(B,R)=(8,2)$.
}
\label{tab:br_detailed_results}

\end{table*}
\begin{table*}[t]
\centering

\footnotesize
\setlength{\tabcolsep}{3pt}
\renewcommand{\arraystretch}{1.12}
\renewcommand{\tabularxcolumn}[1]{m{#1}}

\begin{tabularx}{0.92\textwidth}{
@{}
>{\raggedright\arraybackslash}m{0.13\textwidth}
*{5}{>{\centering\arraybackslash}X}
@{}
}
\toprule

\textbf{Method}
& \shortstack{
    \textbf{Avg. Accuracy}\\
    \textbf{(\%)$\uparrow$}
  }
& \textbf{CFR (\%)$\downarrow$}
& \shortstack{
    \textbf{Training Time}\\
    \textbf{(min/step)}$\downarrow$
  }
& \shortstack{
    \textbf{Response Length}\\
    \textbf{(tokens)}$\downarrow$
  }
& \shortstack{
    \textbf{Inference Time}\\
    \textbf{(min/response)}$\downarrow$
  } \\
\midrule

\multicolumn{6}{c}{
\textit{Student: Qwen3-0.6B-Non-Thinking}
} \\
\midrule

Relay-OPD
& 31.04
& 13.20
& \textbf{4.67}
& \textbf{14,473.80}
& \textbf{0.22} \\

\rowcolor{oursbg}
CD-OPD
& \textbf{32.41}
& \textbf{9.58}
& 5.76
& 14,699.02
& 0.23 \\

\midrule
\multicolumn{6}{c}{
\textit{Student: Qwen3-1.7B-Non-Thinking}
} \\
\midrule

Relay-OPD
& 46.96
& 10.00
& \textbf{4.93}
& \textbf{11,738.32}
& \textbf{0.28} \\

\rowcolor{oursbg}
CD-OPD
& \textbf{48.28}
& \textbf{6.49}
& 6.04
& 11,921.79
& 0.29 \\

\bottomrule
\end{tabularx}
\caption{
Training and inference efficiency of Relay-OPD and CD-OPD
across Qwen3-0.6B and Qwen3-1.7B.
}
\label{tab:efficiency}
\end{table*}

\section{Detailed Results for \texorpdfstring{$B$ and $R$}{B and R}}
\label{app:br_results}

Table~\ref{tab:br_detailed_results} provides the per-benchmark results corresponding to Figure~\ref{Figure5}.
For both evaluated values of $R$, increasing $B$ from 2 to 8 is accompanied by higher accuracy on all eight benchmarks at both student sizes.
The largest gains for Qwen3-0.6B occur on AIME2025, reaching 2.87 and 2.57 percentage points at $R=2$ and $R=4$, respectively.
For Qwen3-1.7B, AIME2026 shows the largest gains, at 2.64 and 2.25 points, respectively.
The advantage of sampling more distinct problems also appears when the total number of student-only responses is held fixed.
Both $(B,R)=(4,2)$ versus $(2,4)$ and $(8,2)$ versus $(4,4)$ favor the configuration with more problems on every benchmark at both student sizes.
For the latter comparison, the per-benchmark gains range from 0.09 to 1.05 percentage points for Qwen3-0.6B and from 0.13 to 0.65 points for Qwen3-1.7B.
Thus, the higher average accuracy reported in the main text reflects improvements across all eight benchmarks rather than gains confined to a subset.
Increasing $R$ also improves the reported accuracy, although the additional gains are small at $B=8$.
Increasing $R$ from 2 to 4 at this value of $B$ yields per-benchmark gains of 0.03--0.33 percentage points for Qwen3-0.6B and 0.00--0.24 points for Qwen3-1.7B.
\textit{Together, these comparisons indicate that CD-OPD benefits across benchmarks from using more distinct problems to compute the task gradient, even when the number of student-only responses is unchanged.}

\section{Training and Inference Efficiency}
\label{app:efficiency}
Table~\ref{tab:efficiency} compares Relay-OPD and CD-OPD in terms of average accuracy across the eight benchmarks, CFR, training time per step, mean student-only response length, and inference time per response.
For Qwen3-0.6B and Qwen3-1.7B, training time increases from 4.67 to 5.76 and from 4.93 to 6.04 minutes per step, respectively.
Meanwhile, CD-OPD improves average accuracy by 1.37 and 1.32 points and reduces CFR by 3.62 and 3.51 points.
During student-only inference, mean response length increases from 14,473.80 to 14,699.02 tokens for Qwen3-0.6B and from 11,738.32 to 11,921.79 tokens for Qwen3-1.7B.
Inference time changes from 0.22 to 0.23 and from 0.28 to 0.29 minutes per response, respectively.
Neither Contextual Digest construction nor online meta-reweighting is required for student-only inference.
\textit{These results show that CD-OPD improves average accuracy and reduces CFR with a modest increase in training time per step, while maintaining student-only response lengths and inference times close to those of Relay-OPD.}

\onecolumn
\begingroup
\setlength{\parindent}{0pt}
\setlength{\parskip}{3pt}

\section{Case Study: Student-Only Reasoning}
\label{app:case_study}

\small
\setlength{\tabcolsep}{0pt}
\setlength{\LTleft}{0pt}
\setlength{\LTright}{0pt}
\setlength{\LTpre}{3pt}
\setlength{\LTpost}{7pt}
\setlength{\LTcapwidth}{\textwidth}
\renewcommand{\arraystretch}{1.10}

\begin{longtable}{
@{}
p{\dimexpr0.5\textwidth-9pt\relax}
@{\hspace{18pt}}
p{\dimexpr0.5\textwidth-9pt\relax}
@{}
}

\label{tab:aime25_case}\\

\toprule

\multicolumn{2}{@{}p{\textwidth}@{}}{%
\textbf{Problem (AIME2025).}
Let $m,n$ be relatively prime positive integers such that
\[
\prod_{k=4}^{63}
\frac{
\log_k\!\left(5^{k^2-1}\right)
}{
\log_{k+1}\!\left(5^{k^2-4}\right)
}
=
\frac{m}{n}.
\]
Find $m+n$.
\hfill
\textbf{Reference answer: 106.}

\smallskip
Both responses are generated by the respective students
independently, without teacher takeover or a Contextual Digest
at inference.
}\\[5pt]

\midrule
\multicolumn{1}{@{}c}{\textbf{Relay-OPD}}
&
\multicolumn{1}{c@{}}{\textbf{CD-OPD}}
\\[3pt]
\midrule
\endfirsthead


\toprule
\multicolumn{1}{@{}c}{\textbf{Relay-OPD}}
&
\multicolumn{1}{c@{}}{\textbf{CD-OPD}}
\\[3pt]
\midrule
\endhead

\midrule
\multicolumn{2}{r@{}}{%
\footnotesize
}\\
\endfoot

\bottomrule
\endlastfoot

\casecell{1. Rewrite and separate the product}{
\casequote{
We can cancel $\log 5$ from numerator and denominator:
}

\[
\frac{(k^2-1)\log(k+1)}
     {(k^2-4)\log k}.
\]

\casequote{
Let's now split the product into two parts:
}

\[
\underbrace{
\prod_{k=4}^{63}\frac{k^2-1}{k^2-4}
}_{\text{rational product}}
\;\cdot\;
\underbrace{
\prod_{k=4}^{63}\frac{\log(k+1)}{\log k}
}_{\text{log product}}.
\]
}
&
\casecell{1. Rewrite and separate the product}{
\casequote{
Now, we can write everything in terms of base $5$.
}

\[
\frac{(k^2-1)\log_5(k+1)}
     {(k^2-4)\log_5 k}.
\]

\casequote{
Let's separate the product into two parts:
factorial-like terms and logarithmic terms.
}

\[
\begin{gathered}
\prod_{k=4}^{63}
\left(
\frac{k-1}{k-2}
\cdot
\frac{k+1}{k+2}
\right)
\\[-1pt]
{}\cdot
\prod_{k=4}^{63}
\frac{\log_5(k+1)}{\log_5 k}.
\end{gathered}
\]
}
\\
\midrule

\casecell{2. Evaluate the two rational factors}{
\casequote{
This is a telescoping product:
}

\[
\prod_{k=4}^{63}\frac{k-1}{k-2}
=
\frac{3}{2}\cdot\frac{4}{3}
\cdots\frac{62}{61}
=
31.
\]

\casequote{
Similarly, look at the second part:
}

\[
\prod_{k=4}^{63}\frac{k+1}{k+2}
=
\frac{5}{6}\cdot\frac{6}{7}
\cdots\frac{64}{65}
=
\casekey{\frac{1}{13}}.
\]

\casequote{
So now, the total product of the terms is:
}

\[
(31)\cdot\left(\frac{1}{13}\right)
=
\casekey{\frac{31}{13}}.
\]

}
&
\casecell{2. Evaluate the two rational factors}{
\casequote{
Wait --- no. Actually:
}

{\footnotesize\color{casegray}
(after revisiting the factorial simplification)}
\par

\[
\prod_{k=4}^{63}\frac{k-1}{k-2}
=
\frac{62}{2}
=
31.
\]

\casequote{
Again, telescoping!
}

\[
\prod_{k=4}^{63}\frac{k+1}{k+2}
=
\frac{5}{6}\cdot\frac{6}{7}
\cdots\frac{64}{65}
=
\casekey{\frac{1}{13}}.
\]

\casequote{
The product of the ratio parts is
}

\[
31\cdot\frac{1}{13}
=
\casekey{\frac{31}{13}}.
\]

}
\\

\pagebreak[4]

\casecell{3. Reassemble the original expression}{
\casequote{
Wait --- but do we have only the first two products?
}

\casequote{
We have now:
}

1. First product:
\[
\prod_{k=4}^{63}\frac{k^2-1}{k^2-4}
=
\casewrong{31}.
\]

{\color{casered}
\textbf{Reading note.}
The earlier factor $1/13$ is no longer included
in the reported rational product.
}
\par

}
&
\casecell{3. Recheck the two rational factors}{
\casequote{
Wait --- let me double-check:
}

\[
\begin{aligned}
\prod_{k=4}^{63}\frac{k-1}{k-2}
&=31,
\\[-2pt]
\prod_{k=4}^{63}\frac{k+1}{k+2}
&=\casekey{\frac{1}{13}}.
\end{aligned}
\]

\casequote{
The product of ratios is
}

\[
31\cdot\frac{1}{13}
=
\casekey{\frac{31}{13}}.
\]

}
\\
\midrule

\casecell{4. Evaluate the logarithmic product}{
\casequote{
This is a telescoping product of ratios.
}

\[
\prod_{k=4}^{63}\frac{\log(k+1)}{\log k}
=
\frac{\log 64}{\log 4}
=
3.
\]
}
&
\casecell{4. Evaluate the logarithmic product}{
\casequote{
So the product telescopes:
}

\[
\prod_{k=4}^{63}
\frac{\log_5(k+1)}{\log_5 k}
=
\frac{\log_5 64}{\log_5 4}.
\]

\casequote{
Wait --- this simplifies exactly!
}

\[
\frac{\log_5 64}{\log_5 4}
=
\frac{6\log_5 2}{2\log_5 2}
=
3.
\]
}
\\
\midrule

\casecell{5. Final recombination and answer}{
\casequote{
Wait --- so both products now are:
}

``1. First product: $31$''\\
``2. Second product: $3$''
\par

\casequote{
Therefore, the total product is:
}

\[
31\cdot3
=
\casewrong{93}.
\]

\casequote{
So $m=93$, $n=1$, $m+n=94$.
}
}
&
\casecell{5. Final recombination and answer}{
\casequote{
So the product is:
}

\[
\frac{31}{13}\cdot3
=
\caseright{\frac{93}{13}}.
\]

\casequote{
So $93/13$ is reduced.
Therefore, $m=93$, $n=13$, and $m+n=106$.
}
}
\\*

\addlinespace[1pt]

\multicolumn{1}{@{}c}{%
$\casewrong{m+n=94}$
\qquad
\textcolor{casered}{\textbf{(incorrect)}}
}
&
\multicolumn{1}{c@{}}{%
$\caseright{m+n=106}$
\qquad
\textcolor{casegreen}{\textbf{(correct)}}
}
\\[7pt]

\end{longtable}


\medskip
\endgroup
\twocolumn

\section{Prompt Templates}
\label{app:prompt_templates}

\subsection{Training and Inference}

Following prior OPD studies~\citep{relay-opd}, we use the same problem-solving instruction for training and evaluation,
including student-only responses for online meta-reweighting.

\begin{cdprompt}{Problem-solving instruction}
[System]
Please reason step by step, and put your
final answer within \boxed{}.

[User]
{problem}
\end{cdprompt}

\subsection{Dependency Graph Construction}

The teacher constructs the dependency graph from the problem,
the response prefix before takeover, and the current teacher guidance.

\begin{cdprompt}{Dependency graph construction}
Construct a directed dependency graph from the
problem, response prefix, and teacher guidance.

Nodes: Include explicitly stated conditions,
variable definitions, intermediate results,
and reasoning steps, including teacher guidance.
Record exact source spans for each node. Merge
spans only if meanings, units, and conditions agree.

Edges: Add i -> j only when i is needed to
interpret or apply j. Use unique node IDs,
valid edge endpoints, and keep the graph acyclic.
If guidance corrects a preceding statement,
use the correction and cite its source. Exclude
superseded claims. Do not add new reasoning.
Return JSON only:
{
  "nodes": [
    {"id": "...", "content": "...",
     "source_spans": [
       {"source": "...", "text": "..."}
     ]}
  ],
  "edges": [{"from": "...", "to": "..."}]
}

Valid sources: problem, response_prefix,
teacher_guidance.

Problem: {problem}
Response prefix: {response_prefix}
Teacher guidance: {teacher_guidance}
\end{cdprompt}

\subsection{Node Summarization}
The teacher summarizes the selected contextual nodes
in dependency order.

\begin{cdprompt}{Node summarization}
Write one concise summary for each selected
contextual node. Use current teacher guidance
to decide which details to retain.

Preserve necessary variable definitions, values,
units, and conditions. Use only source spans and
corrections explicitly stated in the guidance.
Exclude superseded claims. Do not add new
reasoning or rewrite the guidance.

Keep the node IDs and dependency order.
Return one summary per node as JSON only:
{"summaries": [{"id": "...", "text": "..."}]}
Teacher guidance: {teacher_guidance}
Selected nodes: {selected_contextual_nodes}
\end{cdprompt}

The node summaries are concatenated in dependency order,
followed by the unchanged teacher guidance $T$ as the final
Digest segment $D_J=T$.

\end{document}